\documentclass[preprint, 12pt]{elsarticle}
\usepackage{amsmath}

\usepackage{amssymb}
\usepackage{amsthm}
\usepackage{bm}
\let\oldproofname=\proofname
\renewcommand{\proofname}{\rm\bf{\oldproofname}}
\usepackage{ulem}
\usepackage{tikz}
\usetikzlibrary{shadows,arrows}
\pgfdeclarelayer{background}
\pgfdeclarelayer{foreground}
\pgfsetlayers{background,main,foreground}
\usepackage{pgfplots}
\pgfplotsset{compat=1.3}
\usepackage{hyperref}
\usepackage{lineno,hyperref}
\usepackage[export]{adjustbox}
\usepackage{graphicx}
\usepackage{pdfpages}
\usepackage{subcaption}
\usepackage{algorithmic}
\usepackage{algorithm}
\usepackage{multirow}
\usepackage{makecell}
\usepackage{caption} 
\usepackage{booktabs}
\makeatletter
\def\ps@pprintTitle{
\let\@oddhead\@empty
\let\@evenhead\@empty
\let\@oddfoot\@empty
\let\@evenfoot\@oddfoot
}
\makeatother
\usepackage{siunitx}
\usepackage{ wasysym }
\usepackage{ textcomp }

\newdefinition{rmk}{Remark}
\newproof{pf}{Proof}
\newdefinition{defi}{Definition}
\usepackage{tablefootnote}
\begin{document}	
\begin{frontmatter}

\title{Low-rank Dictionary Learning for Unsupervised Feature Selection}
\author[rv1]{Mohsen Ghassemi Parsa}
\ead{mgparsa@ut.ac.ir}
\author[rv1]{Hadi Zare \corref{cor1}}
\ead{h.zare@ut.ac.ir}
\author[rv2]{Mehdi Ghatee}
\ead{ghatee@aut.ac.ir}
\cortext[cor1]{Corresponding author}
\address[rv1]{Faculty of New Sciences and Technologies, University of Tehran, Tehran, Iran}
\address[rv2]{Department of Mathematics and Computer Science, Amirkabir University of Technology, Tehran, Iran}
\begin{abstract}
There exist many high-dimensional data in real-world applications such as biology, computer vision, and social networks.
Feature selection approaches are devised to confront with high-dimensional data challenges with the aim of efficient learning technologies as well as reduction of models complexity.  Due to the hardship of labeling on these datasets, there are a variety of approaches on feature selection process in an unsupervised setting by considering some important characteristics of data. In this paper, we introduce a novel unsupervised feature selection approach by applying dictionary learning ideas in a low-rank representation. Dictionary learning in a low-rank representation not only enables us to provide a new representation, but it also maintains feature correlation. Then, spectral analysis is employed to preserve sample similarities. Finally, a unified objective function for unsupervised feature selection is proposed in a sparse way by an $\ell_{2,1}$-norm regularization. Furthermore, an efficient numerical algorithm is designed to solve the corresponding optimization problem.  We demonstrate the performance of the proposed method based on a variety of standard datasets from different applied domains. Our experimental findings  reveal that the proposed method outperforms the state-of-the-art algorithm.
\end{abstract}

\end{frontmatter}

\section{Introduction}
\label{intro} 
Technological advancement and popularity of social networks provide many huge and high-dimensional data and information sources. 
 High-dimensional data are available in many applications, including machine vision \citep{shi_sparse_2015}, text mining \citep{rogati_high-performing_2002}, and biology \citep{hoseini_unsupervised_2019}. High-dimensionality not only increases the complexity of the training process and the learned model but also degrades performance, that is called as the curse of dimensionality \citep{murphy_machine_2012}. To address the issue, dimensionality reduction can be considered in two main approaches, feature extraction (FE) and feature selection (FS) \citep{pandit_comprehensive_2020}. The new features are constituted by a linear or non-linear transformation of the original features in FE approaches, while FS methods aim to select appropriate features by considering some evaluation criteria. FS attains more attraction than FE in some situations, specifically when the primary aim is to take advantage of more interpretable and understandable features \citep{li_feature_2017}. 

FS methods can be classified to filter, wrapper, and embedded approaches according to feature evaluation. Filters \citep{krzanowski_selection_1987,he_laplacian_2005,zare_relevant_2016} exploit data properties to find out the importance of the features, while wrappers \citep{dy_feature_2004} evaluate the feature subsets by a learning algorithm. In embedded methods \citep{li_unsupervised_2012}, the feature selection process is embedded in a learning algorithm. Recently, unsupervised feature selection (UFS) has attracted many efforts among researchers due to the unavailability of the right answers on practical domains and real-world applications \citep{parsa_unsupervised_2020, zare_similarity_2020}.

UFS algorithms are mainly categorized into similarity preserving, sparse learning, reconstruction, and dictionary learning methods.
Similarity preserving methods \citep{he_laplacian_2005, zhao_spectral_2007} are tried to maintain the local geometric structures among the selected features. 
Sparse learning methods \citep{parsa_unsupervised_2020,zare_similarity_2020} considered selecting more relevant features in a regularized way.
Data reconstruction methods \citep{masaeli_convex_2010,farahat_efficient_2013} re-express the features to eliminate uninformative ones. 
One of the most important approaches in UFS is based on dictionary learning \citep{zhu_coupled_2016, zhu_robust_2017,ding_joint_2020}, in which a new sparse representation of the data matrix is obtained on a dictionary basis space.

Most of the earlier dictionary learning approaches proposed to build the dictionary matrix without any restriction on the rank of the basis matrix. 
A more natural assumption is to employ the low-rank representation on high-dimensional data to alleviate the noisy and redundant features \citep{chen_sparse_2012}. In addition, the basis matrix can be learned in a parsimonious way by imposing the rank constraint, which can be improved the learning process. In this paper, we propose a dictionary learning-based unsupervised feature selection method, named DLUFS, to provide a sparse representation of the original data. Furthermore, we employ a low-rank constraint to eliminate the noisy and redundant features. The local sample structure is also considered by exploiting a spectral analysis.

We summarize the main contributions of this paper as,
\begin{itemize}
	\item A dictionary learning method is proposed to select features to obtain a sparse representation of data.
	\item Low-rank constraint on the basis matrix is imposed to eliminate the noisy and redundant features.
	\item Spectral analysis is employed to preserve the local similarities among samples.	
\end{itemize}

This paper is organized as follows. The existing UFS methods are reviewed in Section \ref{sec:lit_rew}. The proposed method, an illustrative example, and the corresponding optimization algorithm are presented in Section \ref{sec:propos}. We analyze the convergence behavior of the proposed algorithm in Section \ref{sec:conver}. Section \ref{sec:exp} presents the experimental results on benchmark datasets based on state-of-the-art methods. The conclusions are given in Section \ref{sec:concl}.
 
\section{Related Works}\label{sec:lit_rew}
In this section, we review unsupervised feature selection methods in four categories, similarity preserving, sparse machine learning, data reconstruction, and dictionary learning methods.

Similarity preserving methods consider the sample structure in the selected features, such as Laplacian score, LS \citep{he_laplacian_2005}, spectral feature selection, SPEC \citep{zhao_spectral_2007}, and trace ratio criterion for feature selection, TrRatio \citep{nie_trace_2008}. The learning models are not employed in this category of methods which results in the selection of less relevant features.

In sparse machine learning methods, feature selection is performed based on learning a regularized model, such as low-dimensional embedding and sparse regression, JELSR \citep{hou_joint_2014}, local discriminative sparse subspace learning, LDSSL \citep{shang_local_2019}, multi-cluster feature selection, MCFS \citep{cai_unsupervised_2010}, non-negative discriminative feature selection, NDFS \citep{li_unsupervised_2012}, 
similarity preserving feature selection, SPFS \citep{zhao_similarity_2013}, structure preservation robust spectral feature selection, SRFS \citep{zhu_local_2018}, unsupervised discriminative feature selection, UDFS \citep{yang_l21-norm_2011}. These methods commonly select features based on learning a regularized regression matrix without involving data reconstruction.

Data reconstruction approaches were proposed to select features based on their explanation on linear and non-linear transformation of the data, including sparse principal component analysis, CPFS \citep{masaeli_convex_2010}, greedy unsupervised feature selection, GreedyFS \citep{farahat_efficient_2013}, graph regularized feature selection, GRFS \citep{zhao_graph_2016}, embedded reconstruction based unsupervised feature selection, REFS \citep{li_reconstruction-based_2017}, structure preserving unsupervised feature selection, SPUFS \citep{lu_structure_2018}, reconstruction error minimization, REMFS \citep{yang_unsupervised_2019}.

While dictionary learning and reconstruction based methods can be regarded as similar techniques to learn the basis matrix, the dictionary learning methods enable us to provide a new data representation along with the elimination of redundant features. Feature selection process in dictionary learning methods is conducted in two main phases, learning the basis matrix and sparse new data representation. Most of the dictionary learning methods were proposed by considering a two-step procedure such as  \citep{zheng_graph_2011, zhu_robust_2017}.  Graph sparse coding, GSC \citep{zheng_graph_2011} performed dictionary learning and spectral analysis to yield a sparse representation by an $\ell_{1}$-norm regularization. Robust joint graph sparse coding, RJGSC \citep{zhu_robust_2017}, extended GSC by using an $\ell_{2,1}$-norm regularizer. On the other hand, DGL \citep{ding_joint_2020} jointly learns the basis and sparse data matrix in a unified framework. Earlier dictionary-based methods have not constructed the basis matrix by considering the natural low-rank assumption of it, which is justified on many real-world high-dimensional data \citep{chen_sparse_2012}.

\begin{table}[t]
	\centering
	\scriptsize
	\caption{Summary of the state-of-the-art unsupervised feature selection methods.}
	\setlength{\tabcolsep}{4pt}	
	\renewcommand{\arraystretch}{1.2}
	\begin{tabular}{l c c c c c c c}
		\toprule
		{Algorithm}& \makecell{Sparse\\learning} & \makecell{Subspace\\learning} & \makecell{Spectral\\analysis} & \makecell{Joint\\learning} & \makecell{Data\\reconstruction} & \makecell{Dictionary\\learning}& \makecell{Low-rank\\representation} \\ 
		\midrule
		LS \citep{he_laplacian_2005}			&\texttimes&\texttimes&\checkmark&\checkmark&\texttimes&\texttimes&\texttimes\\
		MCFS \citep{cai_unsupervised_2010}	&\checkmark&\checkmark&\checkmark&\texttimes&\texttimes&\texttimes&\texttimes\\
		UDFS \citep{yang_l21-norm_2011}		&\checkmark&\checkmark&\texttimes&\checkmark&\texttimes&\texttimes&\texttimes\\
		NDFS \citep{li_unsupervised_2012}	&\checkmark&\checkmark&\checkmark&\checkmark&\texttimes&\texttimes&\texttimes\\
		SPFS \citep{zhao_similarity_2013}	&\checkmark&\checkmark&\texttimes&\checkmark&\texttimes&\texttimes&\texttimes\\
		JELSR \citep{hou_joint_2014}			&\checkmark&\checkmark&\checkmark&\checkmark&\texttimes&\texttimes&\texttimes\\
		LDSSL \citep{shang_local_2019}		&\checkmark&\checkmark&\checkmark&\checkmark&\checkmark&\texttimes&\texttimes\\
		SRFS \citep{zhu_local_2018} 			&\checkmark&\checkmark&\checkmark&\checkmark&\checkmark&\texttimes&\checkmark\\		
		RJGSC \citep{zhu_robust_2017}		&\checkmark&\texttimes&\checkmark&\texttimes&\checkmark&\checkmark&\texttimes\\
		DGL \citep{ding_joint_2020} 			&\checkmark&\texttimes&\checkmark&\checkmark&\checkmark&\checkmark&\texttimes\\
		DLUFS 								&\checkmark&\checkmark&\checkmark&\checkmark&\checkmark&\checkmark&\checkmark\\		\bottomrule
	\end{tabular}
	\label{tb_comp}	
\end{table}

Table \ref{tb_comp} presents a summary of the related methods by considering important characteristics in a UFS process.
Sparse learning employs a regularization approach to the learning model. Subspace learning indicates the low-dimensional representation of the original data. In spectral analysis, the local structure of the samples is taken into account. 
Joint learning refers to a unified objective function. In data reconstruction, features are expressed by a linear or non-linear combination of all features to discard redundant ones. In dictionary learning, a new representation of the original data is learned in a basis space.
By low-rank representation, the reconstruction matrix is decomposed to low-rank matrices to consider the correlation among features. 
In this paper, we propose a unified UFS method based on all of these main characteristics to yield an efficient and robust procedure.

\section{The Proposed Method}\label{sec:propos}
In this section, at first notations are presented. Then, the proposed method and its algorithm details are introduced. Finally, the proposed algorithm is illustrated through an example.
\subsection{Notations}
In this paper, the vectors and the matrices are denoted by bold lowercase and bold uppercase characters. For a given vector $\mathbf{v}$, its $\ell_{2}$-norm is denoted by $\lVert \mathbf{v} \rVert_2$. Suppose $\mathbf{M}$ is an arbitrary matrix, $M_{ij}$ represents its $(i,j)$-th element, $\mathbf{m_i}$ is the $i$-th row, $\mathbf{m^j}$ is the $j$-th column, $\textrm{tr}(\mathbf{M})$ is the trace, and $\mathbf{M}^{\top}$ is the transpose of the matrix. The Frobenius norm is denoted by $\lVert \mathbf{M} \rVert_F$, and the $\ell_{2,1}$-norm is defined as,
\begin{equation*}
{\lVert \mathbf{M} \rVert}_{2,1}= \sum\limits_{i}\sqrt{\sum\limits_{j}M_{ij}^2}.
\end{equation*}
Let $\mathbf{X} \in \mathbb{R}^{p\times n}$ represents the data matrix, where $p$ is the number of features and $n$ is the number of samples.
\subsection{The Proposed Method}
At first, a new data representation matrix can be learned based on dictionary learning approach as,
\begin{equation}
	\label{eq:12}
	\displaystyle \min_{\mathbf{Q},\mathbf{Z}} {\lVert \mathbf{X} - \mathbf{Q}\mathbf{Z} \rVert}_F^2,
\end{equation}
where  $\mathbf{Z} \in \mathbb{R}^{p\times n}$ is a representation of the data matrix $\mathbf{X}$ in the space of the dictionary matrix $\mathbf{Q} \in \mathbb{R}^{p\times p}$. 
The rank of high-dimensional data can be increased by the noisy and outlier features \citep{chen_sparse_2012}. Based on this fact, the data can be represented in a low-rank space. In this regard, a low-rank constraint on the basis matrix is imposed as,
\begin{equation}
	\label{eq:32}
	\begin{array}{r l} 
		\displaystyle \min_{\mathbf{Q},\mathbf{Z}} & {\lVert \mathbf{X} - \mathbf{Q}\mathbf{Z} \rVert}_F^2\\[-.1em]
		\textrm{s.t.} & \textrm{rank}(\mathbf{Q}) = r,
	\end{array}
\end{equation}
where $r \ll \{n,p\}$ is the induced rank to $\mathbf{Q}$. The low-rank constraint on Eq. \eqref{eq:32} is equivalent to multiply two rank $r$ matrices as,
\begin{equation}
	\label{eq:15}
	\displaystyle \min_{\mathbf{A},\mathbf{B},\mathbf{Z}} {\lVert \mathbf{X} - \mathbf{AB}\mathbf{Z} \rVert}_F^2,
\end{equation}
where $\mathbf{A} \in \mathbb{R}^{p\times r}$, $\mathbf{B} \in \mathbb{R}^{r\times p}$. By $\mathbf{Q} = \mathbf{A}\mathbf{B}$ decomposition, the feature correlation is considered in a low-rank space.
The matrix $\mathbf{Z}$ is transformed to a low-dimensional matrix $\mathbf{BZ} \in \mathbb{R}^{r\times n}$ to perform subspace learning, while $\mathbf{A}$ re-transforms $\mathbf{BZ}$ to the original space. More specifically, as further expressed in Eq. \eqref{eq:11}, the mentioned subspace learning is calculated based on LDA \citep{fukunaga_introduction_1990}. 

The global feature correlations are maintained by low-rank constraint. Furthermore,  the spectral analysis is applied to take the local sample structure into account as,
\begin{equation}
\label{eq:33}
\displaystyle \min_{\mathbf{A},\mathbf{B},\mathbf{Z}} {\lVert \mathbf{X} - \mathbf{AB}\mathbf{Z} \rVert}_F^2 + \alpha\,\textrm{tr}(\mathbf{ZL}\mathbf{Z}^{\top}),
\end{equation}
where $\alpha$ is a tuning parameter. The Laplacian matrix $\mathbf{L}$ is calculated as $\mathbf{L} = \mathbf{D} - \mathbf{S}$, where the diagonal matrix $\mathbf{D}$ is defined as $D_{ii}=\sum_j S_{ij}$ and the similarity matrix $\mathbf{S}$ is calculated as follows,
\begin{equation}
S_{ij} = { 
	\begin{cases}
	\textrm{exp}\left(-\frac{{\lVert \mathbf{x^i} - \mathbf{x^j} \rVert}_2^2}{\sigma^2}\right),& \textrm{if }\ \mathbf{x^i} \in \textrm{N}_k\left(\mathbf{x^j}\right) \textrm{ or } \mathbf{x^j} \in \textrm{N}_k\left(\mathbf{x^i}\right) \\
	0, & \textrm{otherwise},
	\end{cases}}
\end{equation}
where $\textrm{N}_k\left(\mathbf{x^i}\right)$ represents the set of $k$-nearest neighbors of $\mathbf{x^i}$, and $\sigma$ is the width parameter for the Gaussian kernel.

A feature selection framework can be provided by inducing a sparse learning on the new representation matrix $\mathbf{Z}$. Hence, the final objective function is proposed as,
\begin{equation}
	\label{eq:13}
	\displaystyle \min_{\mathbf{A},\mathbf{B},\mathbf{Z}} {\lVert \mathbf{X} - \mathbf{AB}\mathbf{Z} \rVert}_F^2 + \alpha\,\textrm{tr}(\mathbf{ZL}\mathbf{Z}^{\top}) + \lambda\, {\lVert \mathbf{Z} \rVert}_{2,1},
\end{equation}
where $\lambda$ is the regularization parameter. 
The $\ell_{2,1}$-norm regularizer provides sparsity on the rows of  $\mathbf{Z}$, inspired by the group lasso penalty \citep{yuan_model_2006}. The rows are closer to zero, then the corresponding features are more likely regarded as uninformative features.

\subsection{Optimization}
We consider the main objective function as the following optimization problem, 
\begin{equation}
	\label{eq:14} 
	\displaystyle \min_{\mathbf{A},\mathbf{B},\mathbf{Z}} f(\mathbf{A},\mathbf{B},\mathbf{Z})={\lVert \mathbf{X} - \mathbf{AB}\mathbf{Z} \rVert}_F^2 + \alpha\,\textrm{tr}(\mathbf{ZL}\mathbf{Z}^{\top}) + \lambda\, {\lVert \mathbf{Z} \rVert}_{2,1},
\end{equation}
First, by fixing $\mathbf{Z}$ in the main optimization problem in Eq. \eqref{eq:14} to get,
\begin{equation}
\label{eq:9}
\displaystyle \min_{\mathbf{A},\mathbf{B}} f(\mathbf{A},\mathbf{B})={\lVert \mathbf{X} - \mathbf{A}\mathbf{B}\mathbf{Z} \rVert}_F^2.
\end{equation}
By setting the derivative of the Eq. \eqref{eq:9} with respect to $\mathbf{A}$ to zero,
\begin{equation}
\label{eq:7}
\mathbf{A} = \mathbf{X}\mathbf{Z}^{\top}\mathbf{B}^{\top}(\mathbf{B}\mathbf{S_w}\mathbf{B}^{\top})^{-1},
\end{equation}
where $\mathbf{S_w} = \mathbf{Z}\mathbf{Z}^{\top}$ is the correlation among features in the new representation matrix $\mathbf{Z}$.

We rewrite Eq. \eqref{eq:9} as,
\begin{equation}
\label{eq:10}
\begin{array}{r l} 
\displaystyle \min_{\mathbf{A},\mathbf{B}} &\textrm{tr}(\mathbf{X}\mathbf{X}^{\top}) - 2\textrm{tr}(\mathbf{ABZ}\mathbf{X}^{\top}) +\textrm{tr}(\mathbf{ABZ}\mathbf{Z}^{\top}\mathbf{B}^{\top}\mathbf{A}^{\top}).
\end{array}
\end{equation}
Using obtained $\mathbf{A}$ from Eq. \eqref{eq:7} in Eq. \eqref{eq:10} to derive the objective function of $\mathbf{B}$ as,
\begin{align}	
\label{eq:11}
&\displaystyle \min_{\mathbf{B}}\  -\textrm{tr}(\mathbf{X}\mathbf{Z}^{\top}\mathbf{B}^{\top}(\mathbf{B}\mathbf{S_w}\mathbf{B}^{\top})^{-1}\mathbf{BZ}\mathbf{X}^{\top}), \vspace{10pt}\nonumber \\
\Leftrightarrow&\displaystyle \max_{\mathbf{B}}\ \textrm{tr}((\mathbf{B}\mathbf{S_w}\mathbf{B}^{\top})^{-1}\mathbf{BS_b}\mathbf{B}^{\top}),
\end{align}
where $\mathbf{S_b} = \mathbf{Z}\mathbf{X}^{\top}\mathbf{X}\mathbf{Z}^{\top}$. Similar to discriminant analysis \citep{fukunaga_introduction_1990}, $\mathbf{S_w}$ and $\mathbf{S_b}$ can be interpreted as within-class and between-class scatter matrices. Therefore, $\mathbf{B}^{\top}$ can be learned by $r$ eigenvectors of $\mathbf{S_w}^{-1}\mathbf{S_b}$ corresponding to top $r$ eigenvalues.

By rewriting the objective function in Eq. \eqref{eq:13},
\begin{equation}
\label{eq:2}
f(\mathbf{Z}) = {\lVert \mathbf{X} - \mathbf{A}\mathbf{B}\mathbf{Z} \rVert}_F^2 + \alpha\, \textrm{tr}(\mathbf{ZL}\mathbf{Z}^{\top}) + \lambda\, {\lVert \mathbf{Z} \rVert}_{2,1}.
\end{equation}
Let  $\mathbf{A}$ and $\mathbf{B}$ are fixed in Eq. \eqref{eq:2}. Then, by setting the derivative of $f(\mathbf{Z})$ to zero, the following Sylvester equation \citep{bartels_solution_1972} can be obtained,
\begin{equation}
\label{eq:3}
\left((\mathbf{A}\mathbf{B})^{\top}\mathbf{AB}+\lambda \mathbf{D}\right)\mathbf{Z} + \mathbf{Z}\left(\alpha \mathbf{L}\right) = (\mathbf{A}\mathbf{B})^{\top}\mathbf{X},
\end{equation}
where $\mathbf{D}$ is a diagonal matrix as,
\begin{equation}
\label{eq:6}
D_{ii}=\frac{1}{2\lVert \mathbf{z_i} \rVert_{2}+\epsilon}.
\end{equation}
Here, \mbox{Algorithm \ref{alg1}} summarizes the iterative procedure of obtaining the main optimization variables in Eq. \eqref{eq:14}. By descending order of ${\lVert \mathbf{z_{i}} \rVert}_2$'s, the importance of the corresponding features are determined.

\begin{algorithm}[h!] 
	\floatname{algorithm}{Algorithm}
	\renewcommand{\algorithmicrequire}{\textbf{Input:}}
	\renewcommand{\algorithmicensure}{\textbf{Output:}}
	\caption{DLUFS algorithm.} 
	\label{alg1} 
	\begin{algorithmic}[1] 
		\REQUIRE Data matrix $\mathbf{X}\in \mathbb{R}^{p\times n}$ and parameters $\alpha$ and $\lambda$.
		\STATE t = 0.
		\STATE Initialize $\mathbf{Z}^{t} = \mathbf{X}$.
		\REPEAT
		\vspace{4pt}
		\STATE Update $\mathbf{B}^{t+1}$ by solving Eq. \eqref{eq:11}.
		\vspace{4pt}
		\STATE Update $\mathbf{A}^{t+1}$ by Eq. \eqref{eq:7}.
		\vspace{4pt}
		\STATE Update the diagonal matrix $\mathbf{D}^{t+1}$ by Eq. \eqref{eq:6}.
		\vspace{4pt}
		\STATE Update $\mathbf{Z}^{t+1}$ by solving the Sylvester equation in \eqref{eq:3}.
		\vspace{4pt}
		\UNTIL{the convergence of the objective function in \mbox{Eq. \eqref{eq:14}}}.
		\ENSURE Sorting features in descending order of ${\lVert \mathbf{z_{i}} \rVert}_2$'s.
	\end{algorithmic}
\end{algorithm}

\subsection{Computational Complexity}
The computational complexity of \mbox{Algorithm \ref{alg1}} consists of computing $\mathbf{B}$, $\mathbf{A}$ and $\mathbf{Z}$ in each iteration. By considering \eqref{eq:11}, cost of updating $\mathbf{B}$ equals to $\textrm{max}\{\textrm{O}(p^3), \textrm{O}(p^2n)\}$. Next by \eqref{eq:7}, the time complexity of computing $\mathbf{A}$ is $\textrm{max}\{\textrm{O}(p^2n), \textrm{O}(p^2r), \textrm{O}(r^3)\}$. Furthermore,  the time complexity of updating $\mathbf{Z}$ contains two elements, computing the input of Sylvester equation in \eqref{eq:3}, and solving the Sylvester equation, which derives as $\textrm{max}\{\textrm{O}(p^3), \textrm{O}(p^2n), \textrm{O}(p^2r)\}$. Since  the assumption of $r \ll \{p, n\}$ holds on high-dimensional settings, computational complexity  of \mbox{Algorithm \ref{alg1}} reduces to $\textrm{max}\{\textrm{O}(p^3), \textrm{O}(p^2n)\}$.

\subsection{An illustrative example}
\begin{figure}[t]
	\centering
	\includegraphics[width=\textwidth]{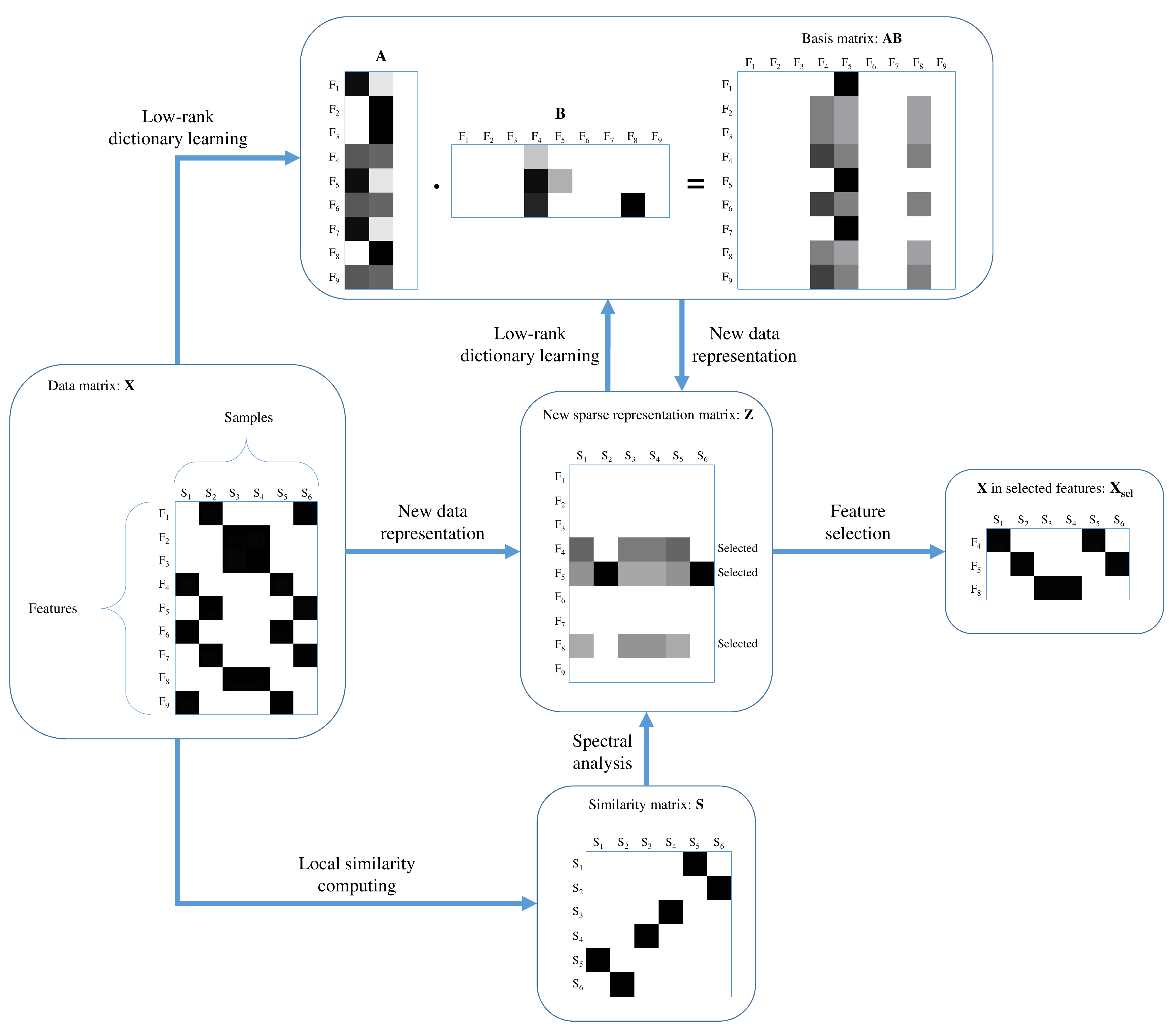}
	\caption{An illustrative example for describing the proposed method.}
	\label{fig:Fig2}
\end{figure}
Fig. \ref{fig:Fig2} describes the proposed algorithm through an illustrative example. All matrices are assumed to be non-negative, where the brighter elements indicate more closeness to zero and the darker ones are far from zero. $\mathbf{X}$ is an artificial data matrix with nine features and six samples. First, a local similarity matrix $\mathbf{S}$ is calculated based on the sample similarities in $\mathbf{X}$. A low-rank basis matrix $\mathbf{AB}$ is obtained by dictionary learning which can be interpreted as the correlation among features based on the low-dimensional matrices $\mathbf{A}$ and $\mathbf{B}$. Then, a new data representation matrix $\mathbf{Z}$ is computed based on spectral analysis in the basis space. The $\mathbf{B}$, $\mathbf{A}$, and $\mathbf{Z}$ are iteratively updated until convergence. Features are ranked by calculating the $\ell_{2}$-norm on their corresponding rows of $\mathbf{Z}$ as our new data matrix. Finally, selected features $\mathbf{X_{sel}}$ are given in the output.

In this example, samples are classified to three sets as $\{\textrm{S}_{1}, \textrm{S}_{5}\}$, $\{\textrm{S}_{2}, \textrm{S}_{6}\}$, and $\{\textrm{S}_{3}, \textrm{S}_{4}\}$ in terms of similarities. In addition, there are three categories of features, $\{\textrm{F}_{1}, \textrm{F}_{5}, \textrm{F}_{7}\}$, $\{\textrm{F}_{2}, \textrm{F}_{3}, \textrm{F}_{8}\}$, and $\{\textrm{F}_{4}, \textrm{F}_{6}, \textrm{F}_{9}\}$ by their similarities. The basis matrix $\mathbf{AB}$ is learned based on low-rank dictionary learning with rank = 3. Fig. \ref{fig:Fig2} indicates that the top three remaining rows of $\mathbf{Z}$ are the $\textrm{F}_{4}$, $\textrm{F}_{5}$, and $\textrm{F}_{8}$. 
Therefore, the output $\mathbf{X_{sel}}$ is formed by selected features. 

\section{Convergence Analysis}
\label{sec:conver}
Our aim is to show the non-increasing behavior of \mbox{Algorithm \ref{alg1}} based on the objective function in Eq. \eqref{eq:14}. Initially, a lemma is given, then the main theorem is presented.
\newtheorem{thm2}{Lemma}
\begin{thm2}\label{lem:1}
	Let $\mathbf{u}$ and $\mathbf{v}$ are two non-zero vectors, then this inequality holds,
	\begin{align}	
	\lVert \mathbf{u} \rVert_{2} - \frac{{\lVert \mathbf{u} \rVert}_{2}^2}{2\lVert \mathbf{v} \rVert_{2}}
	\leq
	\lVert \mathbf{v} \rVert_{2} - \frac{{\lVert \mathbf{v} \rVert}_{2}^2}{2\lVert \mathbf{v} \rVert_{2}}.		
	\end{align}
\end{thm2} 
The proof of Lemma \ref{lem:1} derived in \citep{nie_efficient_2010}.
\newtheorem{thm1}{Theorem}
\begin{thm1}
 Algorithm \ref{alg1} behaves non-increasingly in each update through the primary objective function in \eqref{eq:14}. 
\end{thm1}
\begin{proof}
In the following, the $t$-th iteration of a vector $\mathbf{v}$ and a matrix $\mathbf{M}$ is denoted by $\mathbf{v}^t$ and $\mathbf{M}^t$.

The non-increasing behavior of the objective function of $\mathbf{Z}$ in \eqref{eq:2} is derived by assuming $\mathbf{A}^{t}$ and $\mathbf{B}^{t}$ to be fixed. Since the non-smooth ${\lVert \mathbf{Z} \rVert}_{2,1}$ is iteratively optimized by updating $\mathbf{D}$ and $\mathbf{Z}$, the following inequality can be presented,
\begin{align}
\label{eq:18}
&{\lVert \mathbf{X} - \mathbf{Q}^{t}\mathbf{Z}^{t+1} \rVert}_F^2 + \alpha\, \textrm{tr}(\mathbf{Z}^{t+1}\mathbf{L}{(\mathbf{Z}^{t+1})}^{\top}) + \lambda\, \displaystyle \sum_{i=1}^{p} \frac{{\lVert \mathbf{z_{i}}^{t+1} \rVert}_2^2}{2{\lVert \mathbf{z_{i}}^{t} \rVert}_2} \nonumber\\
\leq
&{\lVert \mathbf{X} - \mathbf{Q}^{t}\mathbf{Z}^{t} \rVert}_F^2 + \alpha\, \textrm{tr}(\mathbf{Z}^{t}\mathbf{L}{(\mathbf{Z}^{t})}^{\top}) + \lambda\, \displaystyle \sum_{i=1}^{p} \frac{{\lVert \mathbf{z_{i}}^{t} \rVert}_2^2}{2{\lVert \mathbf{z_{i}}^{t} \rVert}_2}.
\end{align}
Where $\mathbf{Q} = \mathbf{AB}$.
Then, the inequality \eqref{eq:18} can be rewritten as, 
\begin{align}
	\label{eq:28}
		&{\lVert \mathbf{X} - \mathbf{Q}^{t}\mathbf{Z}^{t+1} \rVert}_F^2 + \alpha\, \textrm{tr}(\mathbf{Z}^{t+1}\mathbf{L}{(\mathbf{Z}^{t+1})}^{\top}) + \lambda\, {\lVert \mathbf{Z}^{t+1} \rVert}_{2,1} - \lambda\, \displaystyle \sum_{i=1}^{p} ({\lVert \mathbf{z_{i}}^{t+1} \rVert}_2 - \frac{{\lVert \mathbf{z_{i}}^{t+1} \rVert}_2^2}{2{\lVert \mathbf{z_{i}}^{t} \rVert}_2})\nonumber\\
		\leq
		&{\lVert \mathbf{X} - \mathbf{Q}^{t}\mathbf{Z}^{t} \rVert}_F^2 + \alpha\, \textrm{tr}(\mathbf{Z}^{t}\mathbf{L}{(\mathbf{Z}^{t})}^{\top}) + \lambda\, {\lVert \mathbf{Z}^{t} \rVert}_{2,1} - \lambda\, \displaystyle \sum_{i=1}^{p} ({\lVert \mathbf{z_{i}}^{t} \rVert}_2 - \frac{{\lVert \mathbf{z_{i}}^{t} \rVert}_2^2}{2{\lVert \mathbf{z_{i}}^{t} \rVert}_2}).
\end{align}
According to Lemma \ref{lem:1}, 
\begin{equation}
\label{eq:19}
{\lVert \mathbf{z_{i}}^{t+1} \rVert}_2 - \frac{{\lVert \mathbf{z_{i}}^{t+1} \rVert}_2^2}{2{\lVert \mathbf{z_{i}}^{t} \rVert}_2}
\leq
{\lVert \mathbf{z_{i}}^{t} \rVert}_2 - \frac{{\lVert \mathbf{z_{i}}^{t} \rVert}_2^2}{2{\lVert \mathbf{z_{i}}^{t} \rVert}_2},
\end{equation}
we obtain,
\begin{align}
\label{eq:20}
&{\lVert \mathbf{X} - \mathbf{Q}^{t}\mathbf{Z}^{t+1} \rVert}_F^2 + \alpha\, \textrm{tr}(\mathbf{Z}^{t+1}\mathbf{L}{(\mathbf{Z}^{t+1})}^{\top}) + \lambda\, {\lVert \mathbf{Z}^{t+1} \rVert}_{2,1}\nonumber \\
\leq
&{\lVert \mathbf{X} - \mathbf{Q}^{t}\mathbf{Z}^{t} \rVert}_F^2 + \alpha\, \textrm{tr}(\mathbf{Z}^{t}\mathbf{L}{(\mathbf{Z}^{t})}^{\top}) + \lambda\, {\lVert \mathbf{Z}^{t} \rVert}_{2,1}.
\end{align}
Therefore,
\begin{equation}
	\label{eq:17}
	f(\mathbf{A}^{t},\mathbf{B}^{t},\mathbf{Z}^{t+1}) \leq f(\mathbf{A}^{t},\mathbf{B}^{t},\mathbf{Z}^{t}).
\end{equation}

In the same way, by fixing $\mathbf{Z}^{t+1}$, it can be shown that,
\begin{equation}
	\label{eq:21}
	f(\mathbf{A}^{t+1},\mathbf{B}^{t+1},\mathbf{Z}^{t+1}) \leq f(\mathbf{A}^{t},\mathbf{B}^{t},\mathbf{Z}^{t+1}).
\end{equation}
By considering inequalities \eqref{eq:17} and \eqref{eq:21},
\begin{equation}
\label{eq:22}
f(\mathbf{A}^{t+1},\mathbf{B}^{t+1},\mathbf{Z}^{t+1}) \leq f(\mathbf{A}^{t+1},\mathbf{B}^{t+1},\mathbf{Z}^{t}) \leq f(\mathbf{A}^{t},\mathbf{B}^{t},\mathbf{Z}^{t}).
\end{equation}
Therefore, the non-increasing behavior of Algorithm \ref{alg1} based on the primary objective function in Eq. \eqref{eq:14} is given.
\end{proof}

\section{Experiments}\label{sec:exp}
This section is divided into five subsections to describe our experimental setup. A brief summary of the applied datasets in our study, evaluation measures, the details of parameters setting, the obtained results, and the sensitivity analysis are presented in the following.
\subsection{Datasets}
A variety of domains of applications are considered in employed datasets such as images of digits (BA \citep{belhumeur_eigenfaces_1997}), colon cancer (Colon \citep{alon_broad_1999}), malignant brain tumor (GLIOMA \citep{c_l_nutt_et_al_gene_2003}), non-sparse artificial dataset (Madelon \citep{li_feature_2017}), image of faces (ORL, Yale \citep{cai_orthogonal_2006}, and WarpAR10P \citep{li_feature_2017}), and PC versus MAC from 20-newsgroups dataset (PCMAC \citep{lang_newsweeder_1995}).
The BA is available on \url{https://cs.nyu.edu/~roweis/data.html}, while all others are accessed from \citep{li_feature_2017}. \mbox{Table \ref{tb_datasets}} reports the main characteristics of datasets. 

\begin{table}[t]
	\centering
	\footnotesize
	\caption{The statistics of datasets.}
	\begin{tabular}{ l c c c c c}
		\toprule
		Dataset & samples & features & classes & Type & Category\\
		\midrule 
		BA & 1404 & 320 & 36 & Binary & Image\\
		Colon & 62 & 2000 & 2 & Discrete & Biology \\
		GLIOMA & 50 & 4434 & 4 & Continuous & Biology \\
		Madelon & 2600 & 500 & 2 & Continuous & Artificial \\
		ORL & 400 & 1024 & 40 & Discrete & Image \\	
		PCMAC&1943&3289&2&Discrete&Text\\
		WarpAR10P&130&2400&10&Discrete& Image \\
		Yale & 165 & 1024 & 15 &Discrete& Image \\
		\bottomrule
	\end{tabular}\\
	\label{tb_datasets}
\end{table}

\subsection{Evaluation measures}\label{subsec:eval_cri}
The clustering techniques are usually applied to evaluate the UFS methods \citep{li_feature_2017}.
Based on the attained clustering results and the ground truth information, two common evaluation measures are used frequently,
Accuracy and Normalized Mutual Information.

Let the ground truth and the predicted one based on clustering approach are shown by $\mathbf{y}$ and $\mathbf{\hat{y}}$. \\
The Accuracy (called as ACC) is defined as,
\begin{equation*}
	\textrm{ACC}(\mathbf{y},\mathbf{\hat{y}})=\frac{1}{n} \sum_{i=1}^{n} \delta(y_i,\textrm{map}(z_i)),
\end{equation*}
where the function $\delta(a,b)$ equals to 1, when $a=b$, and 0 elsewhere. For the $\textrm{map}(.)$ function, the Kuhn-Munkres approach \citep{lovasz_matching_1986} is employed to find the best permutation for matching the categories in vectors $\mathbf{y}$ and $\mathbf{\hat{y}}$.\\
Based on definitions of entropy measure $\textrm{H}(.)$  and the mutual information of $\mathbf{y}$ and $\mathbf{\hat{y}}$ denoted by $\textrm{I}(\mathbf{y},\mathbf{\hat{y}})$ \citep{bishop_pattern_2006}, Normalized Mutual Information (called as NMI)  is given as,
\begin{equation*}
	\textrm{NMI}(\mathbf{y},\mathbf{\hat{y}}) = \frac{\textrm{I}(\mathbf{y},\mathbf{\hat{y}})}{\max(\textrm{H}(\mathbf{y}),\textrm{H}(\mathbf{\hat{y}}))},
\end{equation*}

\subsection{The experimental setting}
We compare the proposed method, DLUFS, with the state-of-the-art unsupervised feature selection algorithms, including JELSR \citep{hou_joint_2014}, LDSSL \citep{shang_local_2019},
 LS \citep{he_laplacian_2005}, 
 MCFS \citep{cai_unsupervised_2010}, NDFS\citep{li_unsupervised_2012}, 
 SPFS \citep{zhao_similarity_2013}, SRFS \citep{zhu_local_2018}, 
 UDFS \citep{yang_l21-norm_2011}, and selecting all features namely Baseline.

The number of neighborhoods in k-nearest neighbor algorithm is set to $k=5$. We set $\sigma = 1$ in our method and others requiring a similarity matrix based on a Gaussian kernel.
For NDFS method, the default $\gamma = 10^8$ is considered. 
Finding the suitable choices of  the tuning parameters $\alpha$ and $\lambda$ in our method is performed by a grid search approach from the set of $\{10^{-4}, 10^{-2}, 1, 10^2, 10^4\}$ candidates, where a similar approach is used to set the tuning parameters in the other methods. The stopping convergence condition for all of the iterative algorithms, including our method, is given as $\frac{\lvert \textrm{f}(t)-\textrm{f}(t-1) \rvert}{\textrm{f}(t)}<10^{-3}$, where $\textrm{f}(t)$ equals to the objective function in the $t$-th iteration.
The NMI and ACC measures are reported on 20 times repetitions. In each repetition, the k-means algorithm ia applied by setting the numbers of features from $\{50, 100, 150, 200, 250, 300\}$. We report two main descriptive statistical quantities of NMI and ACC  in these repeated experiments, mean and standard deviation (STD).

\begin{table}[t]
	\centering
	\caption{The results of clustering (ACC\% $\pm$ std) of UFS methods on standard datasets. The best and the second-best are presented in bold and underlined numbers.}
	\tiny
	\setlength{\tabcolsep}{3.5pt}	
	\renewcommand{\arraystretch}{1.2}
	\label{tab:acc_res}
	\begin{tabular}
		{
			l
			S[table-format=2.2]@{\,\(\pm\)\,}
			S[table-format=1.2]
			S[table-format=2.2]@{\,\(\pm\)\,}
			S[table-format=1.2]
			S[table-format=2.2]@{\,\(\pm\)\,}
			S[table-format=1.2]
			S[table-format=2.2]@{\,\(\pm\)\,}
			S[table-format=1.2]
			S[table-format=2.2]@{\,\(\pm\)\,}
			S[table-format=1.2]
			S[table-format=2.2]@{\,\(\pm\)\,}
			S[table-format=1.2]
			S[table-format=2.2]@{\,\(\pm\)\,}
			S[table-format=1.2]
			S[table-format=2.2]@{\,\(\pm\)\,}
			S[table-format=1.2]
			S[table-format=2.2]@{\,\(\pm\)\,}
			S[table-format=1.2]
		}
		\toprule
		Dataset & \multicolumn{2}{c}{BA} & \multicolumn{2}{c}{Colon} & \multicolumn{2}{c}{GLIOMA} & \multicolumn{2}{c}{Madelon}& \multicolumn{2}{c}{ORL}& \multicolumn{2}{c}{PCMAC}& \multicolumn{2}{c}{WarpAR10P}& \multicolumn{2}{c}{Yale} \\
		\midrule	
		Baseline&{\textbf{43.04}}&{1.18}&{54.84}&{0.00}&{\uline{61.30}}&{4.11}&{50.30}&{0.07}&{\textbf{59.24}}&{2.17}&{50.54}&{0.04}&{21.04}&{2.93}&{41.91}&{2.36}\\
		JELSR&{39.97}&{2.14}&{56.76}&{1.34}&{52.53}&{1.00}&{\uline{57.53}}&{1.21}&{57.22}&{3.15}&{50.55}&{0.03}&{\uline{33.96}}&{1.50}&{37.56}&{1.05}\\
		LDSSL&{40.73}&{3.28}&{58.01}&{0.47}&{56.98}&{0.70}&{52.23}&{1.57}&{43.09}&{5.51}&{\uline{50.69}}&{0.07}&{33.60}&{0.51}&{40.23}&{0.85}\\
		LS&{41.59}&{3.47}&{57.80}&{0.60}&{55.75}&{2.28}&{50.35}&{0.04}&{48.13}&{2.92}&{50.45}&{0.03}&{32.60}&{3.43}&{\uline{44.07}}&{2.42}\\
		MCFS&{41.75}&{2.16}&{54.66}&{1.44}&{60.55}&{4.73}&{52.30}&{0.99}&{57.72}&{1.27}&{50.46}&{0.14}&{23.38}&{3.39}&{39.76}&{0.49}\\
		NDFS&{39.42}&{4.06}&{57.39}&{1.45}&{57.37}&{2.14}&{57.06}&{1.64}&{53.43}&{2.73}&{50.59}&{0.03}&{31.38}&{1.14}&{37.15}&{1.47}\\
		SPFS&{41.39}&{2.07}&{58.33}&{1.61}&{55.68}&{4.48}&{51.58}&{0.12}&{58.08}&{1.28}&{50.52}&{0.04}&{30.96}&{5.55}&{41.33}&{0.71}\\
		SRFS&{37.78}&{4.73}&{\uline{60.32}}&{0.92}&{61.20}&{1.32}&{52.00}&{1.65}&{58.17}&{1.24}&{50.54}&{0.02}&{31.19}&{2.61}&{42.06}&{1.88}\\
		UDFS&{40.69}&{3.05}&{55.67}&{0.86}&{54.17}&{3.05}&{57.47}&{1.66}&{54.74}&{2.10}&{50.57}&{0.05}&{28.60}&{3.78}&{33.53}&{1.33}\\
		DLUFS&{\uline{42.38}}&{1.91}&{\textbf{64.83}}&{7.83}&{\textbf{64.22}}&{2.18}&{\textbf{59.13}}&{0.43}&{\uline{59.22}}&{1.25}&{\textbf{50.80}}&{0.12}&{\textbf{35.65}}&{2.23}&{\textbf{47.31}}&{0.98}\\
		\bottomrule
	\end{tabular}
\end{table}

\begin{table}[t]
	\centering
	\caption{The results of clustering (NMI\% $\pm$ std) of UFS methods on standard datasets. The best and the second-best are presented in bold and underlined numbers.}
	\tiny
	\setlength{\tabcolsep}{3.5pt}	
	\renewcommand{\arraystretch}{1.2}
	\label{tab:nmi_res}
	\begin{tabular}
		{
			l
			S[table-format=2.2]@{\,\(\pm\)\,}
			S[table-format=1.2]
			S[table-format=2.2]@{\,\(\pm\)\,}
			S[table-format=1.2]
			S[table-format=2.2]@{\,\(\pm\)\,}
			S[table-format=1.2]
			S[table-format=2.2]@{\,\(\pm\)\,}
			S[table-format=1.2]
			S[table-format=2.2]@{\,\(\pm\)\,}
			S[table-format=1.2]
			S[table-format=2.2]@{\,\(\pm\)\,}
			S[table-format=1.2]
			S[table-format=2.2]@{\,\(\pm\)\,}
			S[table-format=1.2]
			S[table-format=2.2]@{\,\(\pm\)\,}
			S[table-format=1.2]
			S[table-format=2.2]@{\,\(\pm\)\,}
			S[table-format=1.2]
		}
		\toprule
		Dataset & \multicolumn{2}{c}{BA} & \multicolumn{2}{c}{Colon} & \multicolumn{2}{c}{GLIOMA} & \multicolumn{2}{c}{Madelon}& \multicolumn{2}{c}{ORL}& \multicolumn{2}{c}{PCMAC}& \multicolumn{2}{c}{WarpAR10P}& \multicolumn{2}{c}{Yale} \\
		\midrule	
		Baseline&{\textbf{58.21}}&{0.69}&{00.62}&{0.00}&{\uline{50.93}}&{2.47}&{00.00}&{0.00}&{\uline{77.81}}&{0.83}&{00.04}&{0.04}&{17.41}&{3.60}&{48.93}&{1.85}\\
		JELSR&{55.62}&{1.60}&{02.76}&{0.86}&{31.11}&{2.10}&{01.60}&{0.43}&{75.70}&{1.94}&{00.90}&{0.05}&{33.85}&{2.17}&{44.90}&{1.13}\\
		LDSSL&{56.04}&{3.26}&{01.52}&{0.30}&{49.74}&{0.37}&{00.22}&{0.20}&{66.24}&{3.99}&{01.08}&{0.12}&{\uline{35.00}}&{1.17}&{47.08}&{0.80}\\
		LS&{57.14}&{3.02}&{01.80}&{0.20}&{49.77}&{2.14}&{00.00}&{0.00}&{71.32}&{2.13}&{\uline{01.23}}&{0.30}&{33.10}&{4.42}&{\uline{49.83}}&{2.15}\\
		MCFS&{56.64}&{1.58}&{00.23}&{0.13}&{37.64}&{7.80}&{00.18}&{0.13}&{76.78}&{0.61}&{00.89}&{0.37}&{20.30}&{4.59}&{47.10}&{0.68}\\
		NDFS&{54.65}&{4.23}&{01.71}&{1.40}&{49.87}&{0.46}&{01.51}&{0.54}&{73.43}&{2.00}&{00.67}&{0.07}&{29.73}&{1.19}&{44.38}&{1.31}\\
		SPFS&{56.91}&{1.62}&{01.21}&{0.48}&{35.41}&{6.24}&{00.07}&{0.01}&{77.22}&{0.77}&{00.90}&{0.06}&{27.09}&{6.04}&{48.24}&{0.76}\\
		SRFS&{53.44}&{4.68}&{\uline{02.96}}&{0.89}&{49.62}&{0.59}&{00.21}&{0.32}&{77.27}&{0.84}&{00.09}&{0.09}&{29.16}&{1.37}&{49.60}&{1.10}\\
		UDFS&{56.25}&{2.72}&{01.13}&{0.26}&{29.13}&{4.92}&{\uline{02.13}}&{1.14}&{74.87}&{1.52}&{00.83}&{0.26}&{25.43}&{3.63}&{42.00}&{1.00}\\
		DLUFS&{\uline{57.63}}&{1.84}&{\textbf{07.59}}&{9.66}&{\textbf{51.75}}&{0.34}&{\textbf{02.43}}&{0.23}&{\textbf{78.02}}&{0.71}&{\textbf{02.24}}&{0.48}&{\textbf{36.65}}&{1.07}&{\textbf{56.20}}&{0.62}\\		
		\bottomrule
	\end{tabular}
\end{table}

\begin{figure}[!t]
	\centering
	\captionsetup[subfigure]{font=scriptsize,labelfont=scriptsize}
	\begin{subfigure}[b]{.32\textwidth}
		\includegraphics[scale=0.5]{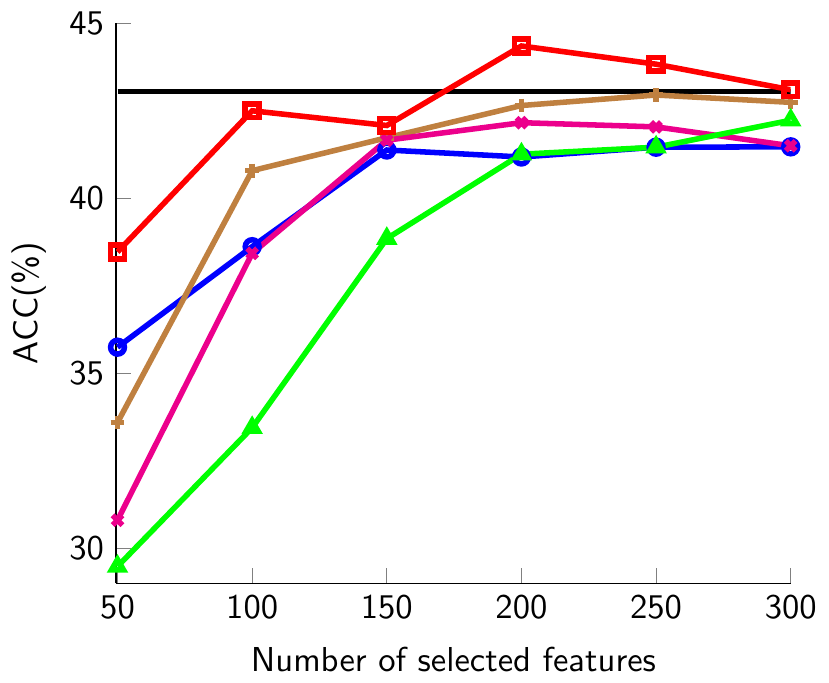}
		\vspace{3pt}
		\caption{BA}
		\label{fig:acc_BA}
	\end{subfigure}
	\bigskip
	\begin{subfigure}[b]{.32\textwidth}
		\includegraphics[scale=0.5]{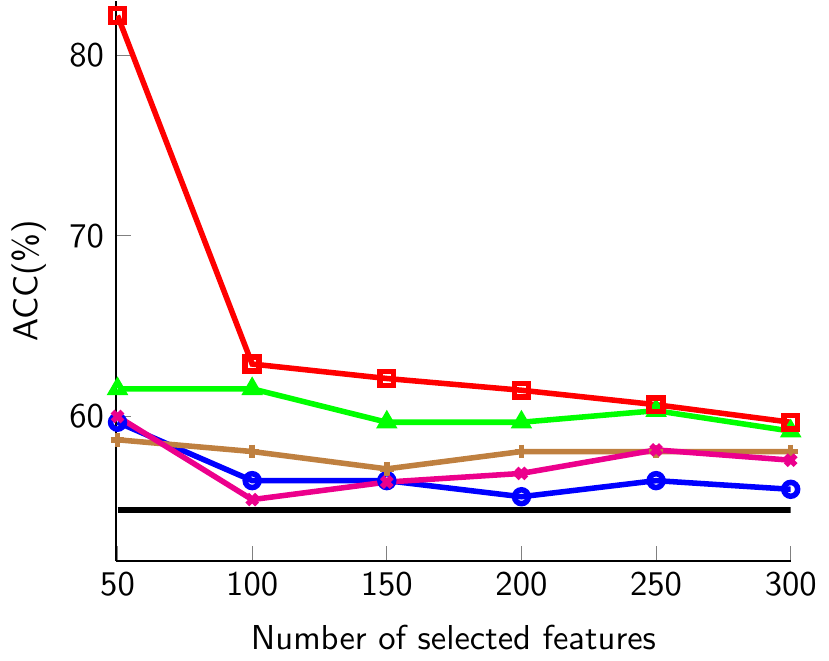}
		\vspace{3pt}
		\caption{Colon}
		\label{fig:acc_Colon}
	\end{subfigure}
	\begin{subfigure}[b]{.32\textwidth}
		\includegraphics[scale=0.5]{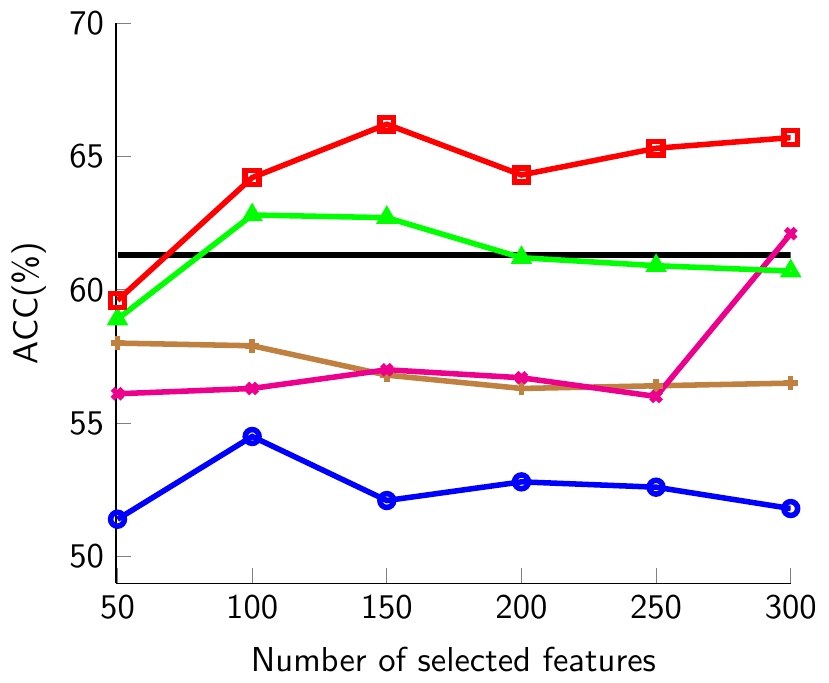}
		\vspace{3pt}
		\caption{GLIOMA}
		\label{fig:acc_GLIOMA}
	\end{subfigure}
	\begin{subfigure}[b]{.32\textwidth}
		\includegraphics[scale=0.5]{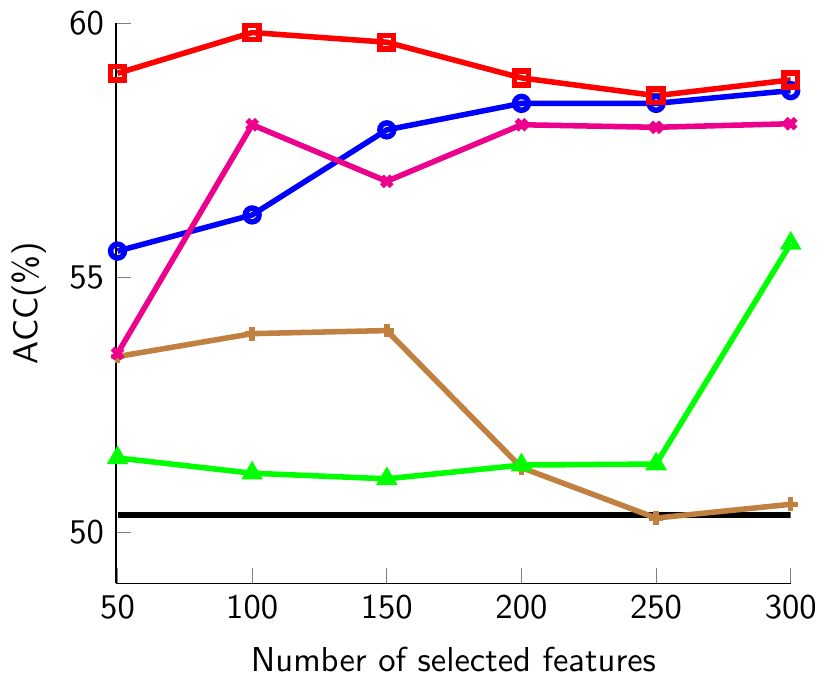}
		\vspace{3pt}
		\caption{Madelon}
		\label{fig:acc_Madelon}
	\end{subfigure}
	\bigskip
	\begin{subfigure}[b]{.32\textwidth}
		\includegraphics[scale=0.5]{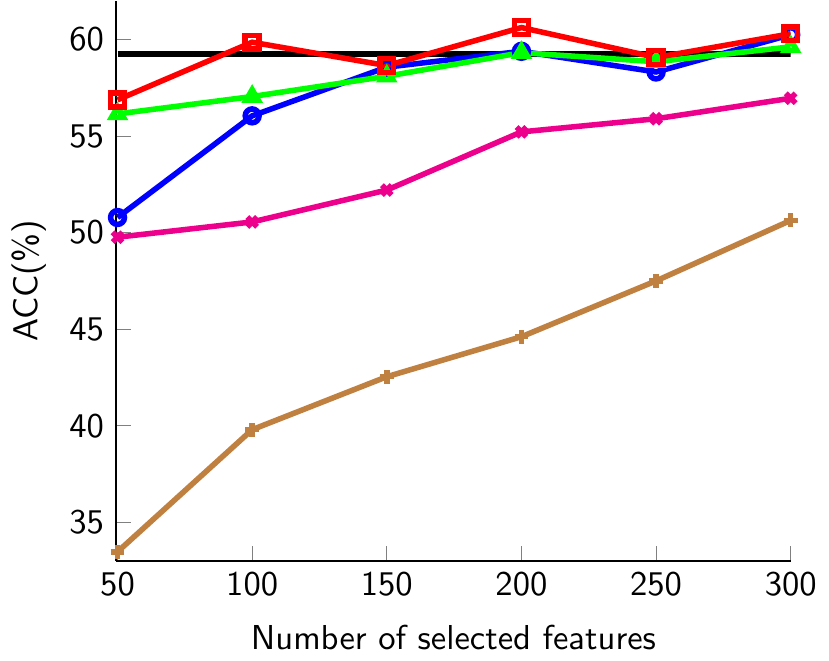}
		\caption{ORL}
		\label{fig:acc_ORL}
	\end{subfigure}
	\begin{subfigure}[b]{.32\textwidth}
		\includegraphics[scale=0.5]{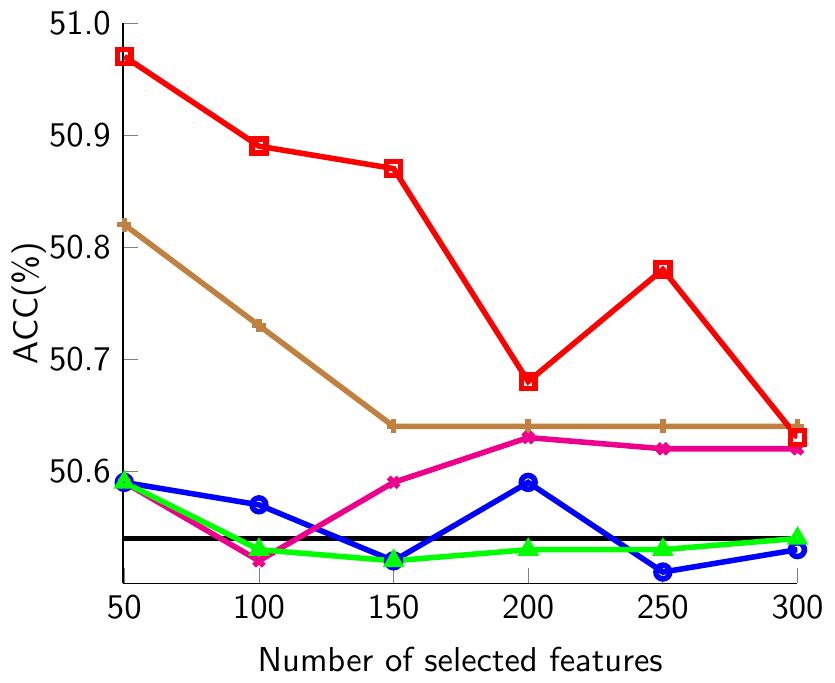}
		\vspace{3pt}
		\caption{PCMAC}
		\label{fig:acc_PCMAC}
	\end{subfigure}
	\begin{subfigure}[b]{.32\textwidth}
		\includegraphics[scale=0.5]{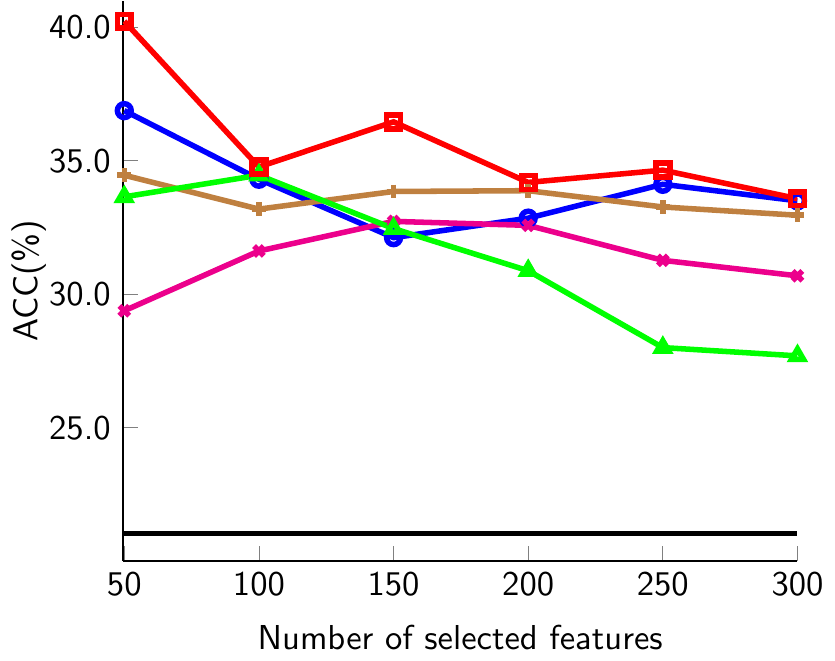}
		\vspace{3pt}
		\caption{WarpAR10P}
		\label{fig:acc_warpAR10P}
	\end{subfigure}
	\bigskip
	\begin{subfigure}[b]{.32\textwidth}
		\includegraphics[scale=0.5]{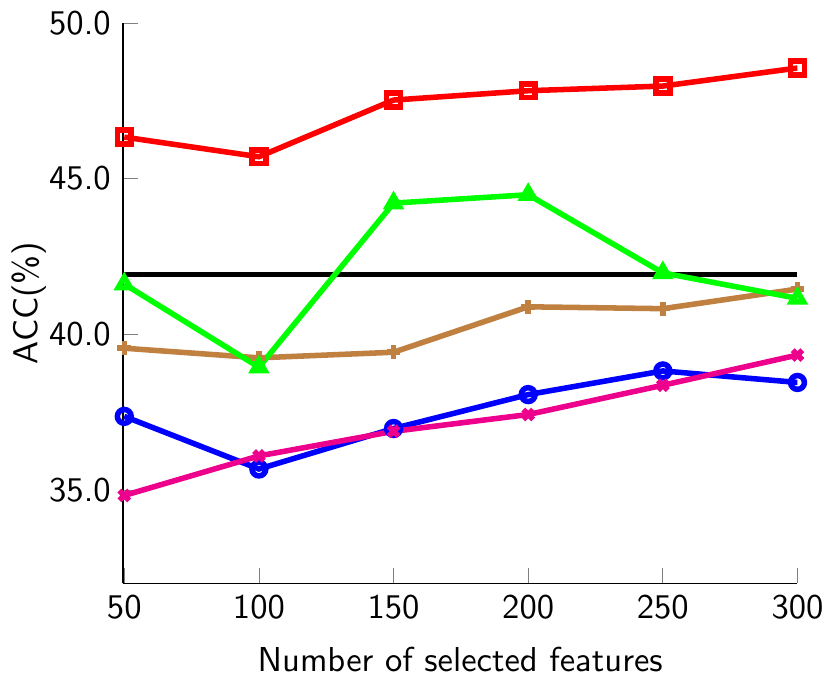}
		\vspace{3pt}
		\caption{Yale}
		\label{fig:acc_Yale}
	\end{subfigure}
	\begin{subfigure}[b]{.32\textwidth}
		\centering
		\includegraphics[valign=t,scale=1.1]{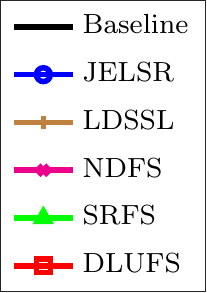}
		\vspace{3pt}
	\end{subfigure}
	\caption{The achieved results in ACC measure by selecting different numbers of features.}
	\label{fig:acc_linechart1}
\end{figure}

\begin{figure}[!h]
	\centering
	\captionsetup[subfigure]{font=scriptsize,labelfont=scriptsize}
	\begin{subfigure}[b]{.32\textwidth}
		\includegraphics[scale=0.5]{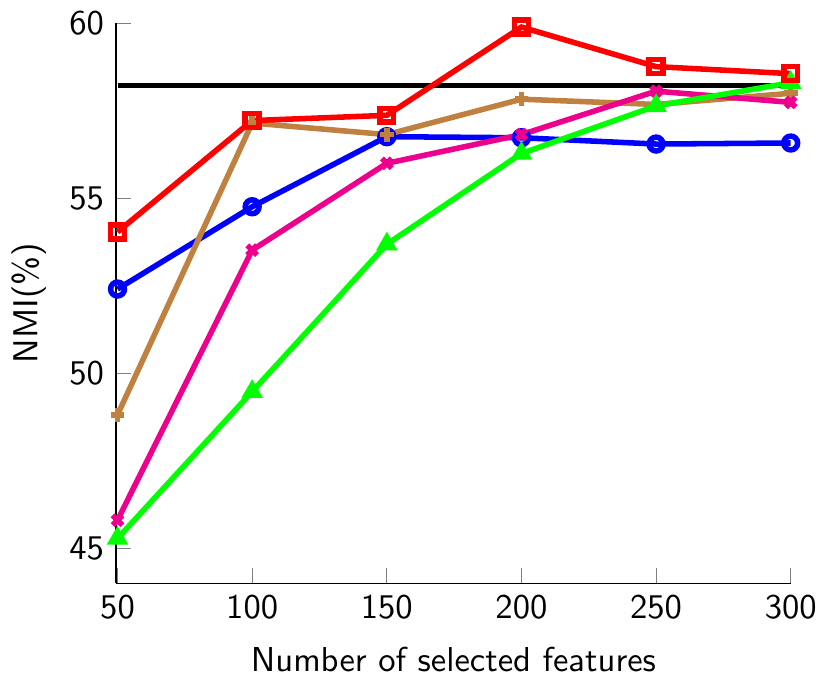}
		\vspace{3pt}
		\caption{BA}
		\label{fig:NMI_BA}
	\end{subfigure}
	\bigskip
	\begin{subfigure}[b]{.32\textwidth}
		\includegraphics[scale=0.5]{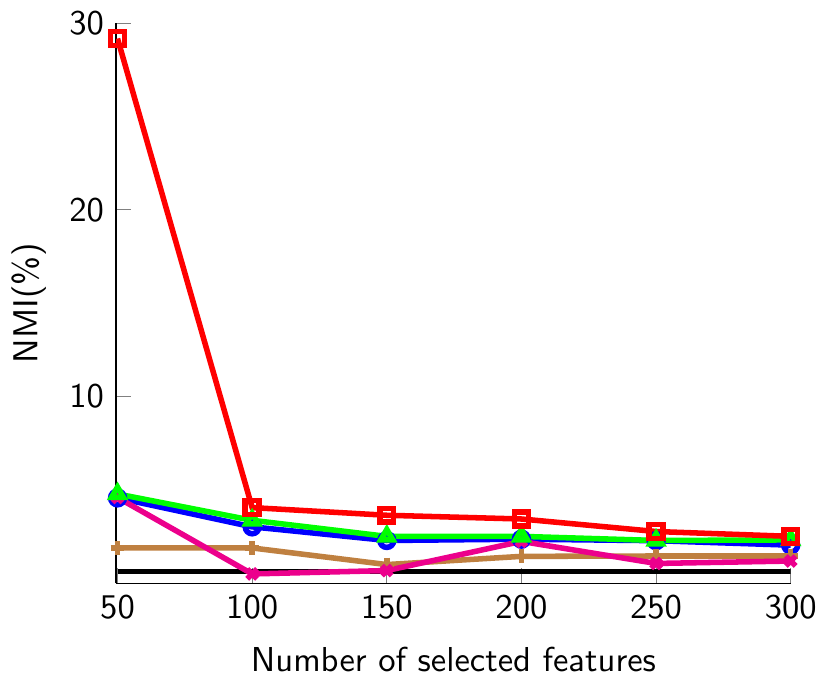}
		\vspace{3pt}
		\caption{Colon}
		\label{fig:NMI_Colon}
	\end{subfigure}
	\begin{subfigure}[b]{.32\textwidth}
		\includegraphics[scale=0.5]{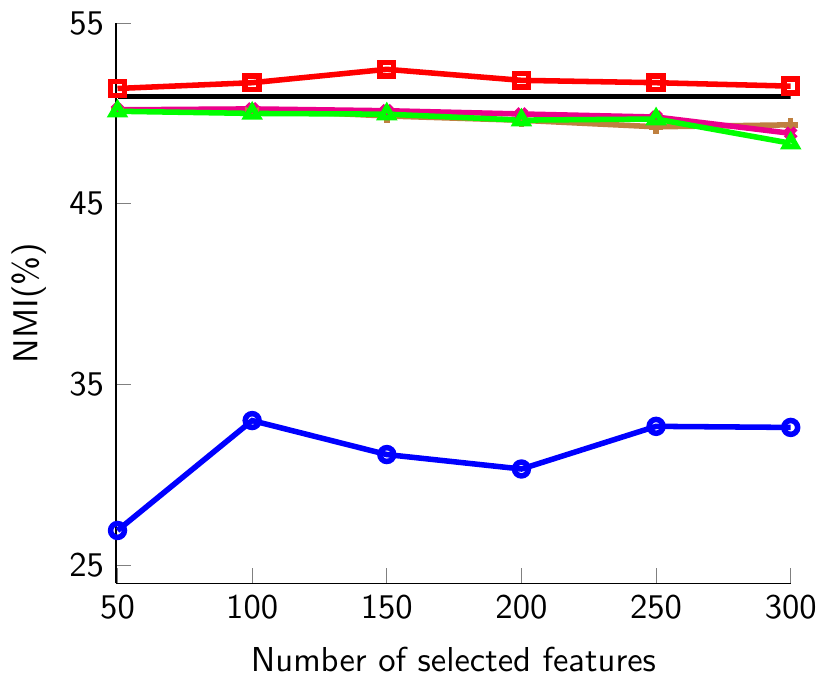}
		\vspace{3pt}
		\caption{GLIOMA}
		\label{fig:NMI_GLIOMA}
	\end{subfigure}
	\begin{subfigure}[b]{.32\textwidth}
		\includegraphics[scale=0.5]{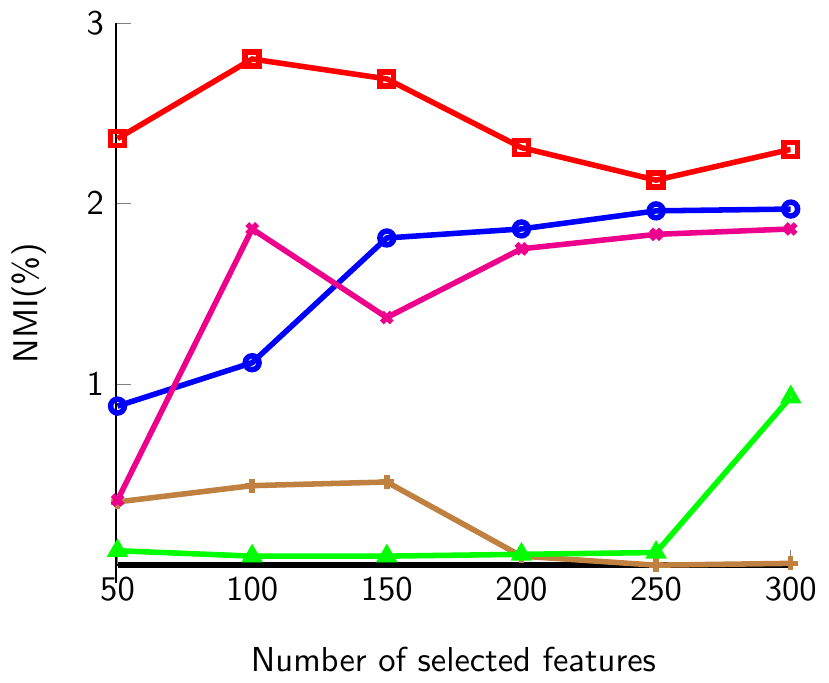}
		\vspace{3pt}
		\caption{Madelon}
		\label{fig:NMI_Madelon}
	\end{subfigure}
	\bigskip
	\begin{subfigure}[b]{.32\textwidth}
		\includegraphics[scale=0.5]{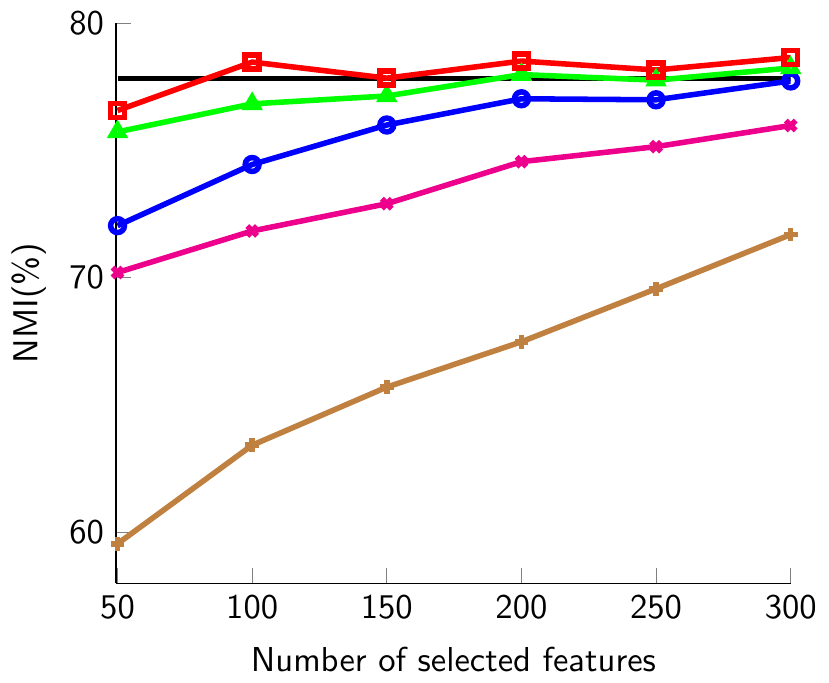}
		\caption{ORL}
		\label{fig:NMI_ORL}
	\end{subfigure}
	\begin{subfigure}[b]{.32\textwidth}
		\includegraphics[scale=0.5]{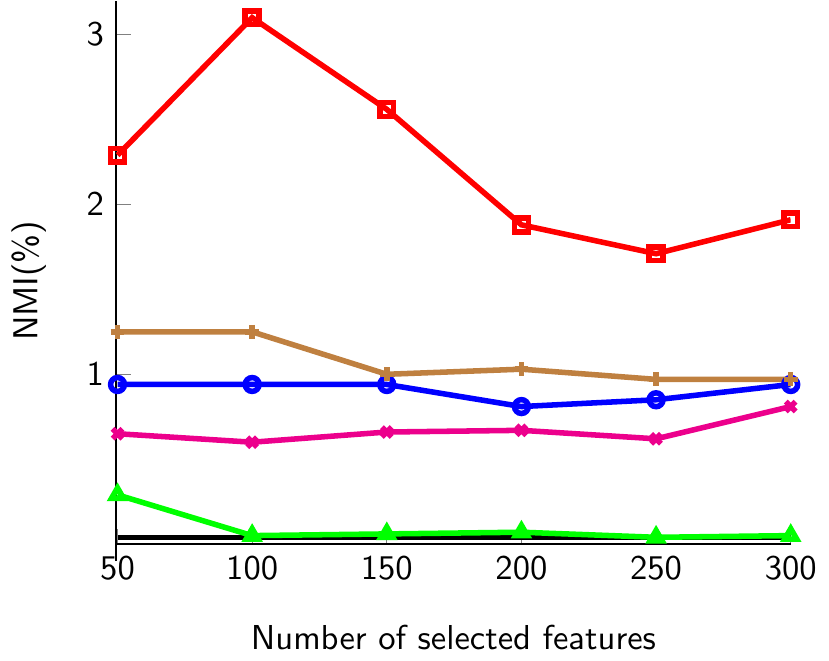}
		\vspace{3pt}
		\caption{PCMAC}
		\label{fig:NMI_PCMAC}
	\end{subfigure}
	\begin{subfigure}[b]{.32\textwidth}
		\includegraphics[scale=0.5]{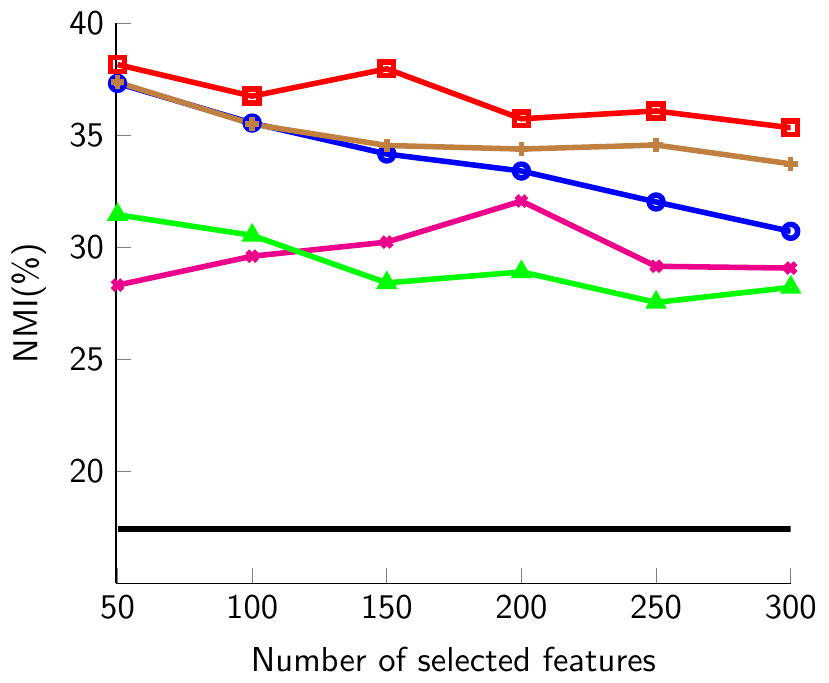}
		\vspace{3pt}
		\caption{WarpAR10P}
		\label{fig:NMI_warpAR10P}
	\end{subfigure}
	\bigskip
	\begin{subfigure}[b]{.32\textwidth}
		\includegraphics[scale=0.5]{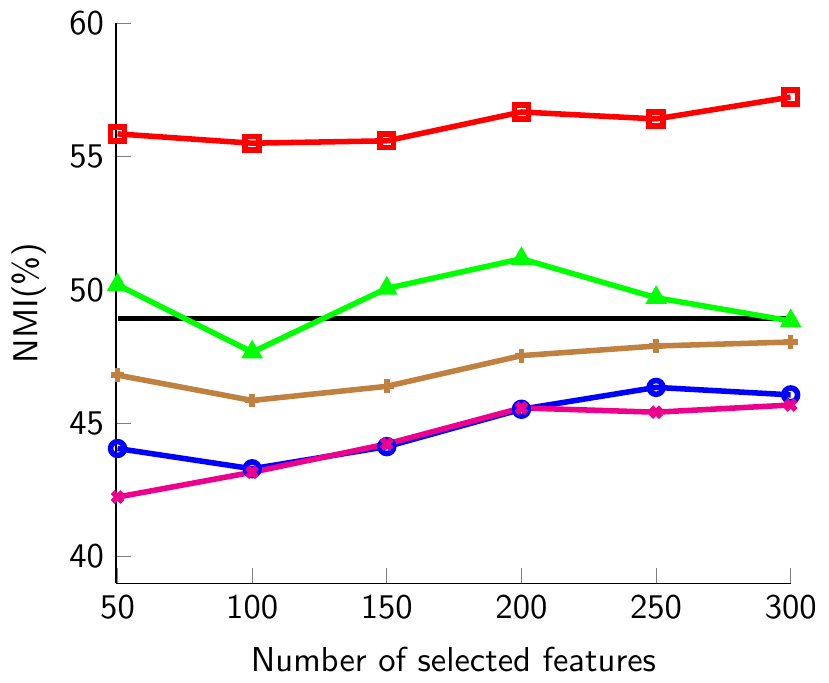}
		\vspace{3pt}
		\caption{Yale}
		\label{fig:NMI_Yale}
	\end{subfigure}
	\begin{subfigure}[b]{.32\textwidth}
		\centering
		\includegraphics[valign=t,scale=1.1]{Label.pdf}
		\vspace{3pt}
	\end{subfigure}
	\caption{The achieved results in NMI measure by selecting different numbers of features.}
	\label{fig:NMI_linechart1}
\end{figure}

\begin{figure}[!h]
	\captionsetup[subfigure]{font=scriptsize,labelfont=scriptsize}
	\centering
	\begin{subfigure}[b]{0.32\textwidth}
		\includegraphics[width=\textwidth]{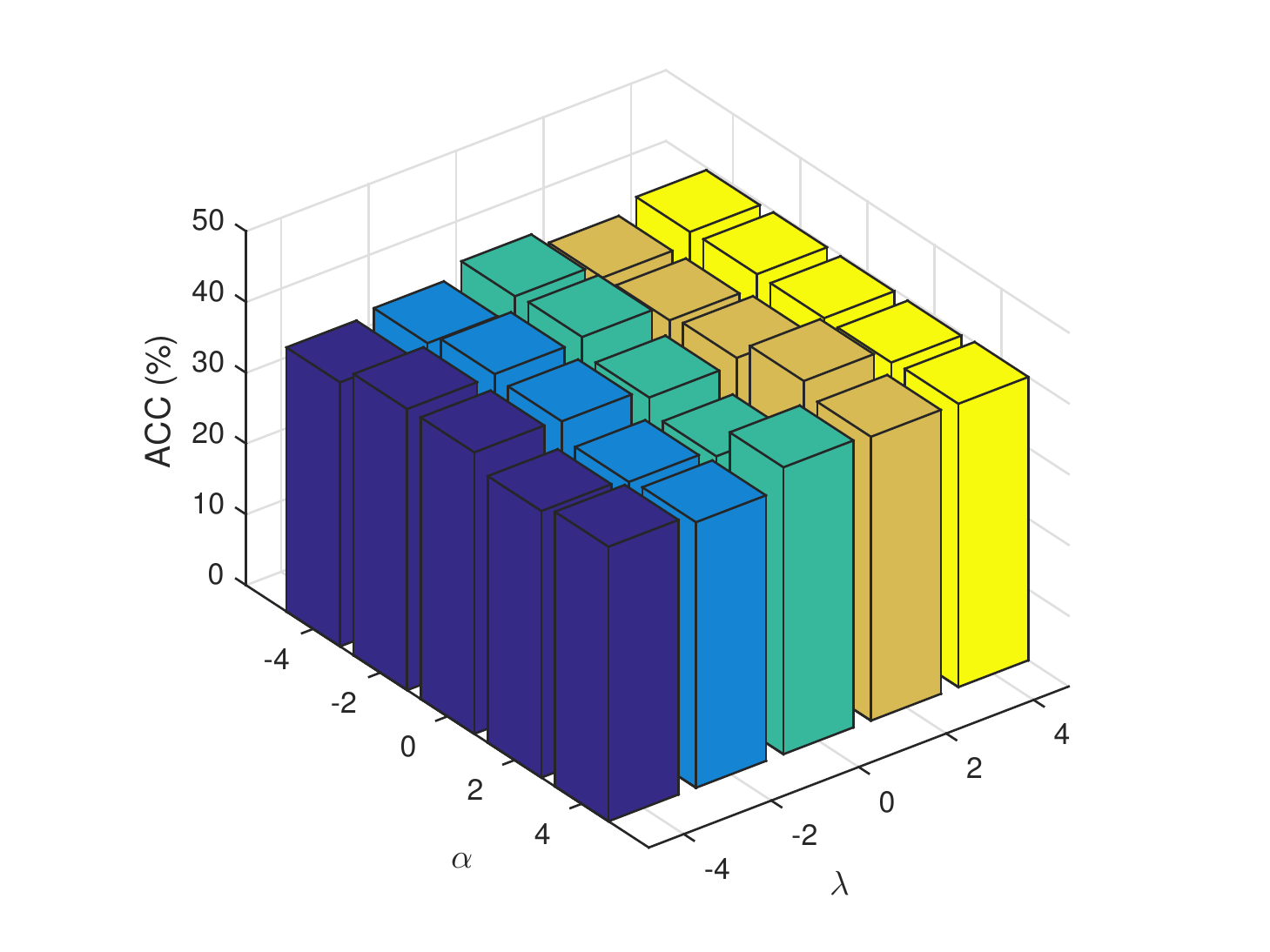}
		\caption{BA}
		\label{fig:BA_acc}
	\end{subfigure}\vspace{1mm}
	\begin{subfigure}[b]{0.32\textwidth}
		\includegraphics[width=\textwidth]{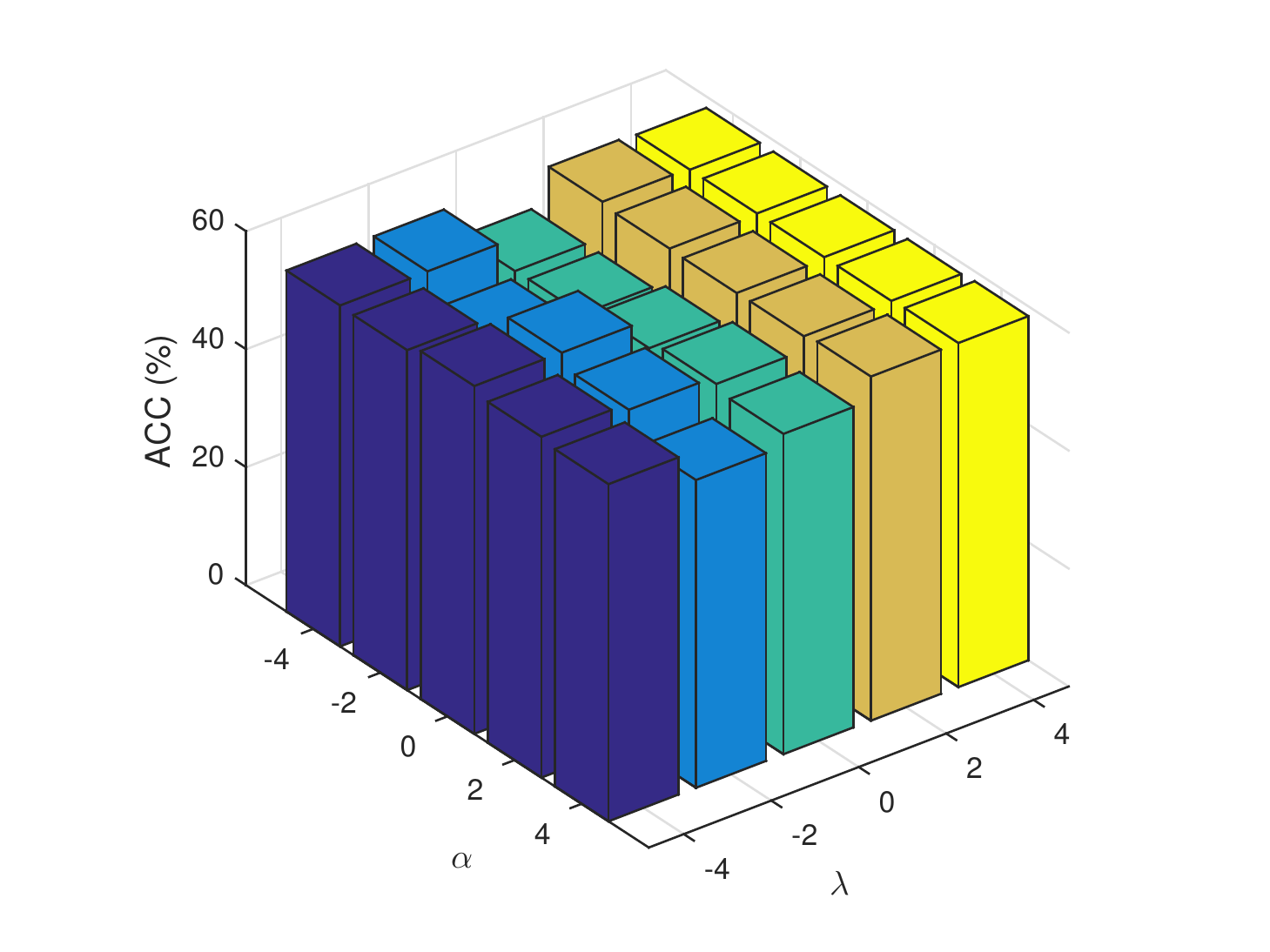}
		\caption{Colon}
		\label{fig:Colon_acc}
	\end{subfigure}\vspace{1mm}
	\begin{subfigure}[b]{0.32\textwidth}
		\includegraphics[width=\textwidth]{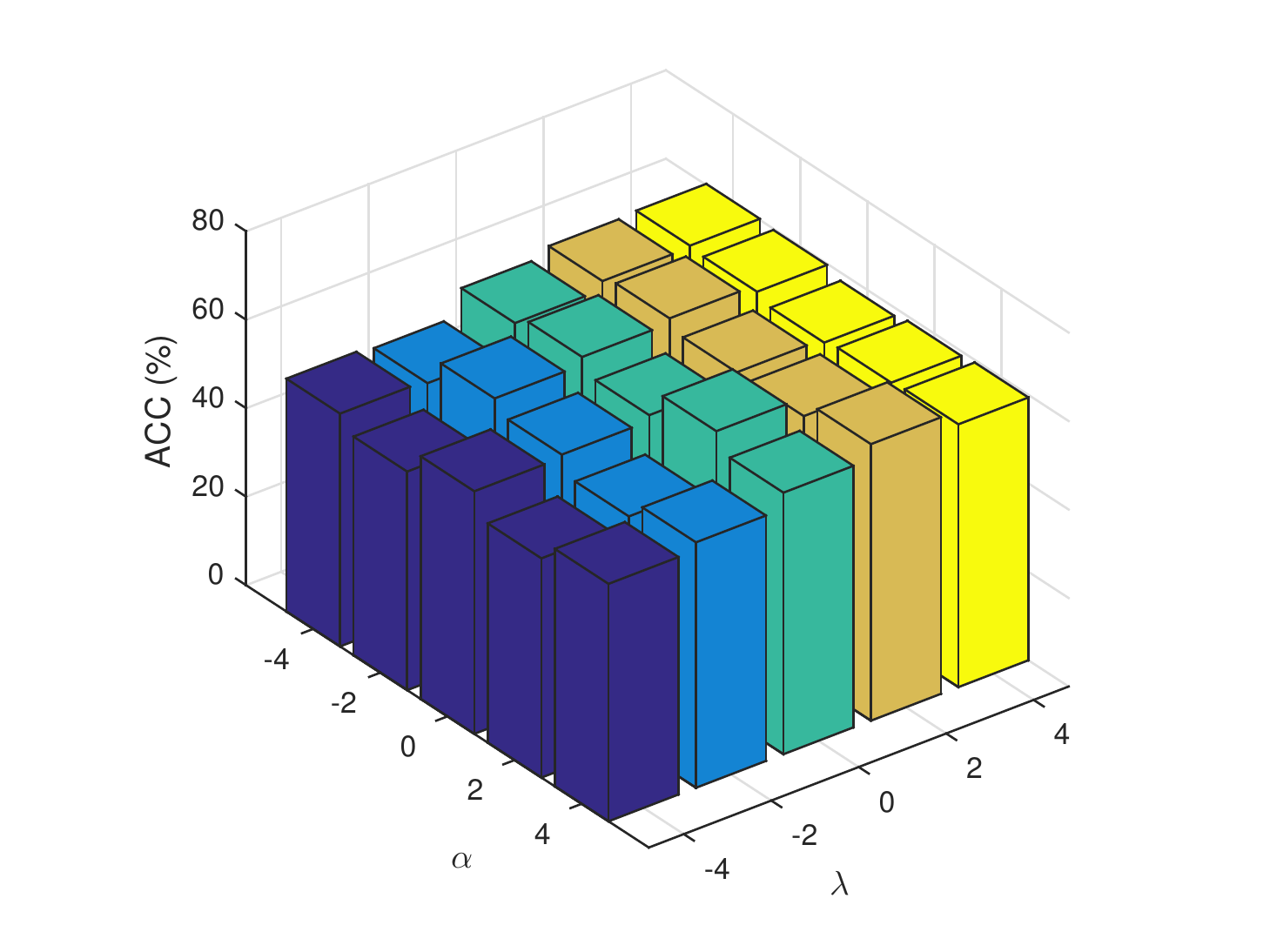}
		\caption{GLIOMA}
		\label{fig:GLIOMA_acc}
	\end{subfigure}\vspace{1mm}
	\begin{subfigure}[b]{0.32\textwidth}
		\includegraphics[width=\textwidth]{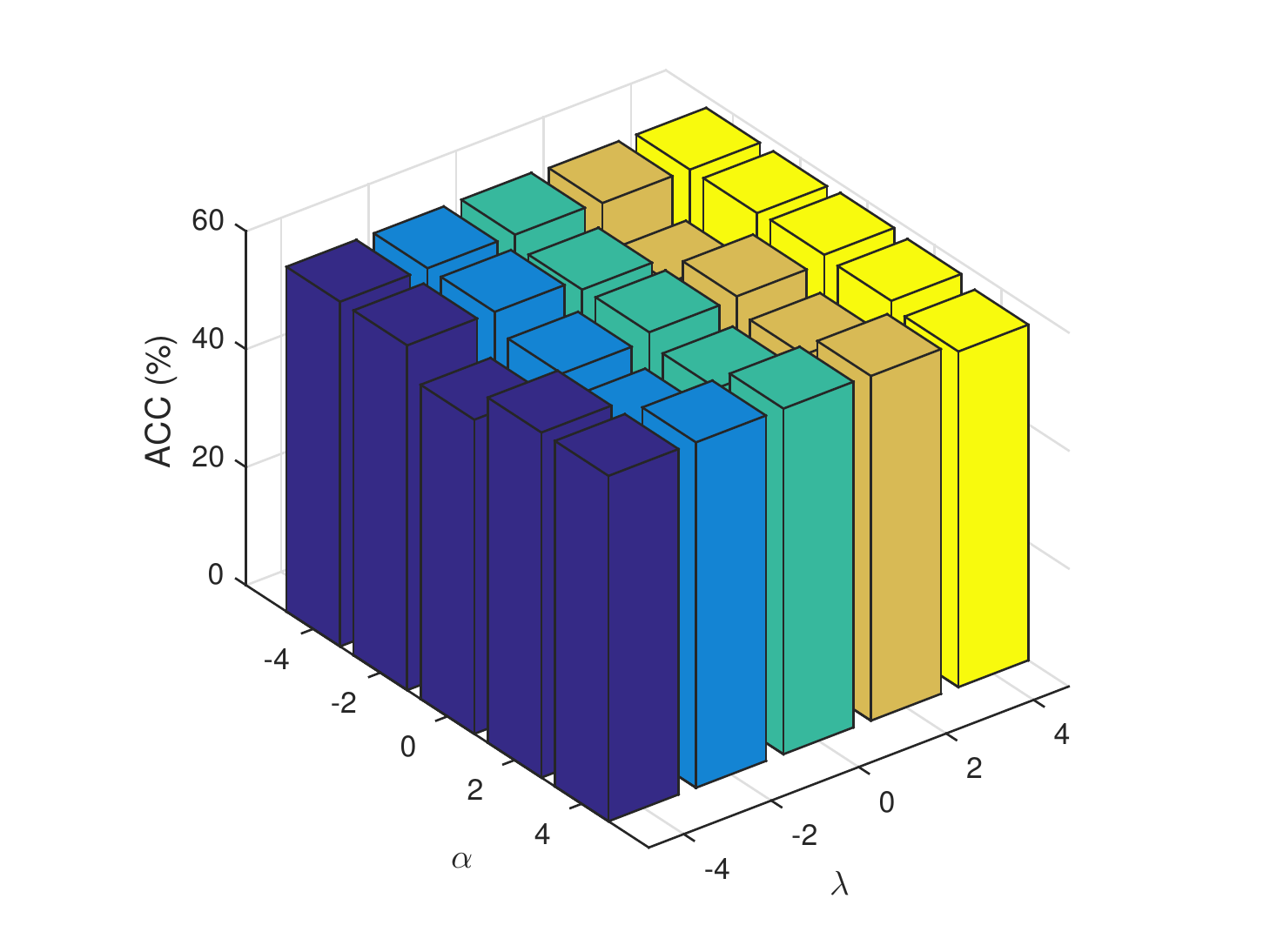}
		\caption{Madelon}
		\label{fig:Madelon_acc}
	\end{subfigure}\vspace{1mm}
	\begin{subfigure}[b]{0.32\textwidth}
		\includegraphics[width=\textwidth]{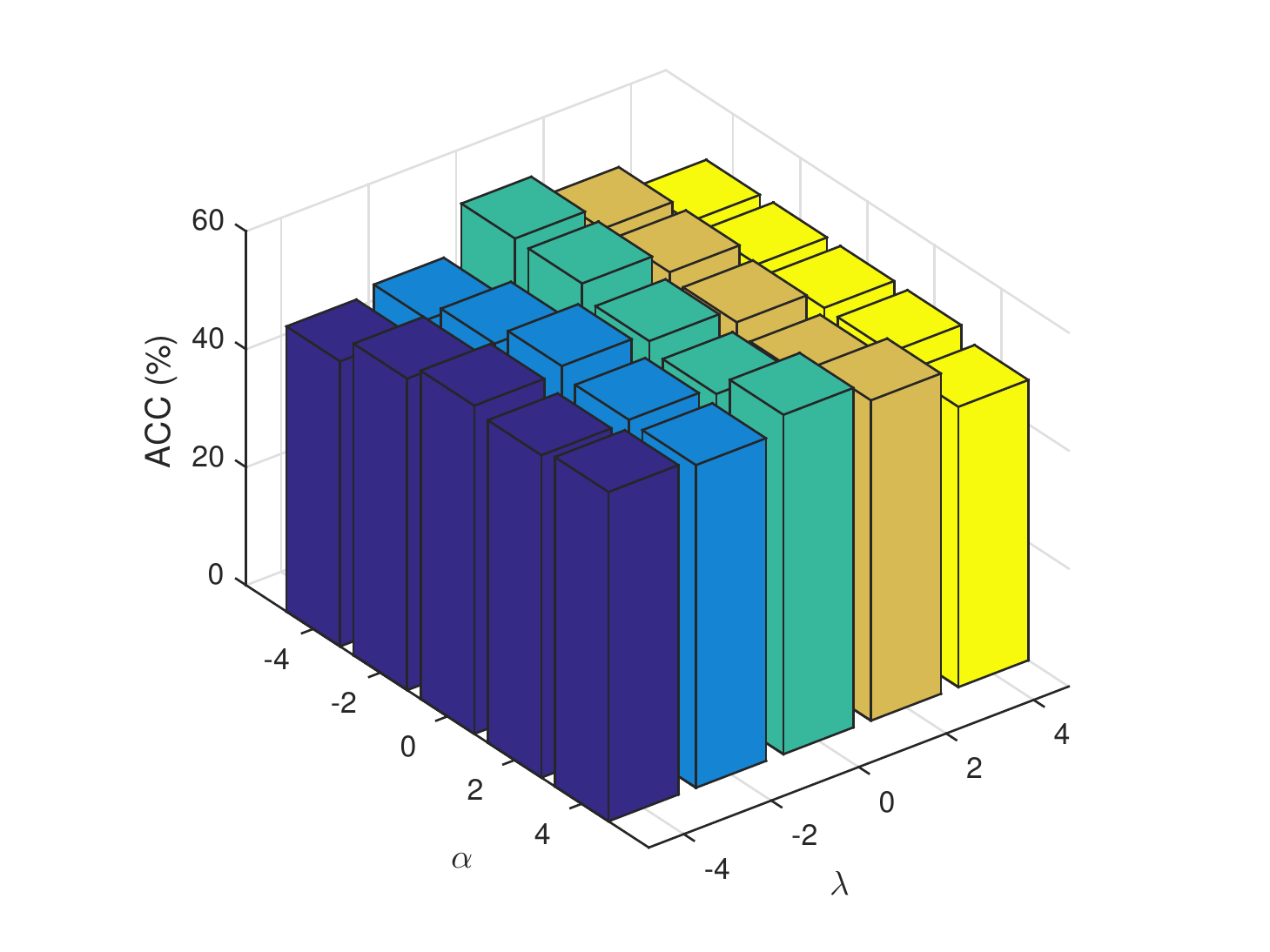}
		\caption{ORL}
		\label{fig:ORL_acc}
	\end{subfigure}\vspace{1mm}
	\begin{subfigure}[b]{0.32\textwidth}
		\includegraphics[width=\textwidth]{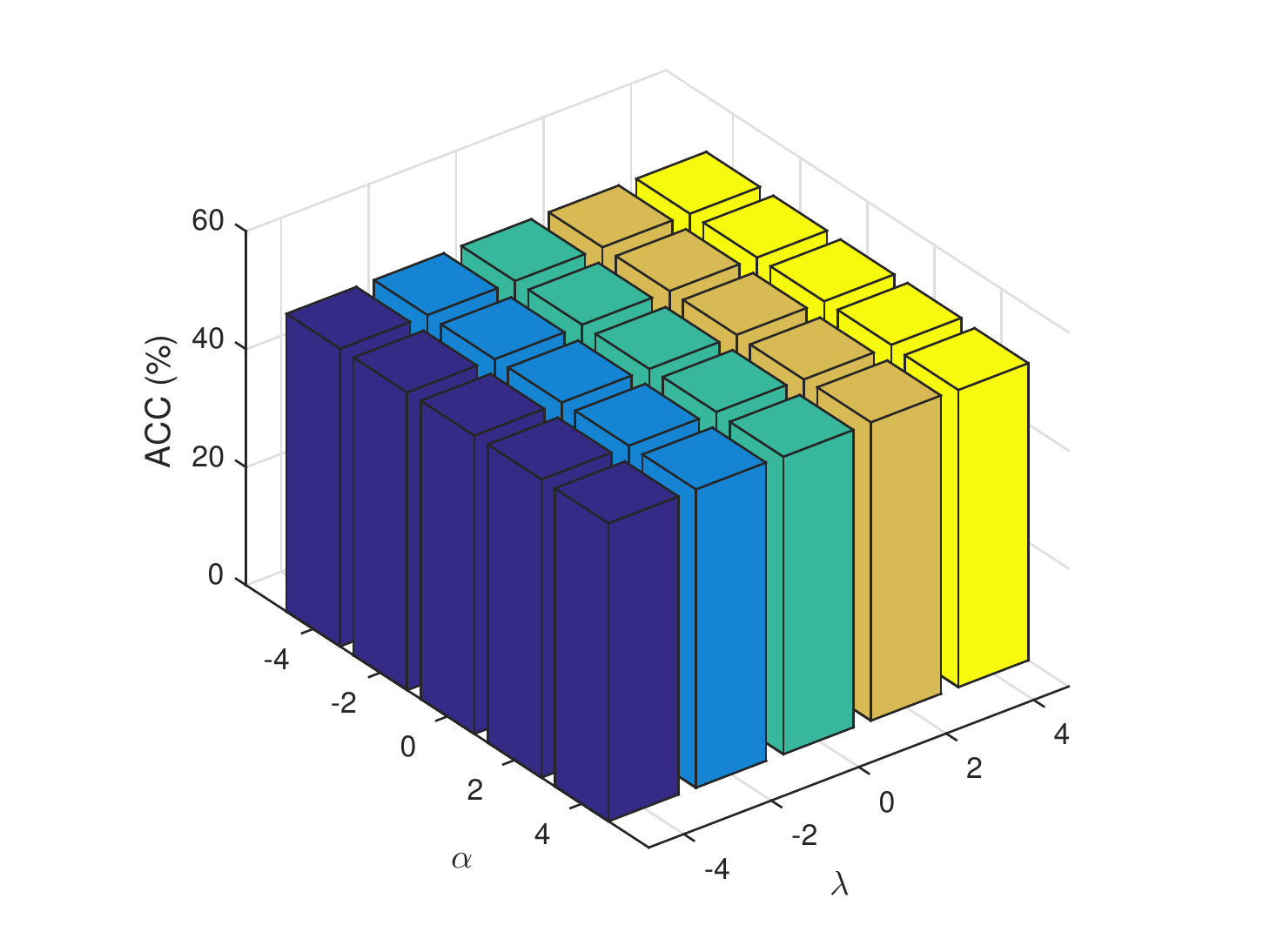}
		\caption{PCMAC}
		\label{fig:PCMAC_acc}
	\end{subfigure}\vspace{1mm}
	\begin{subfigure}[b]{0.32\textwidth}
		\includegraphics[width=\textwidth]{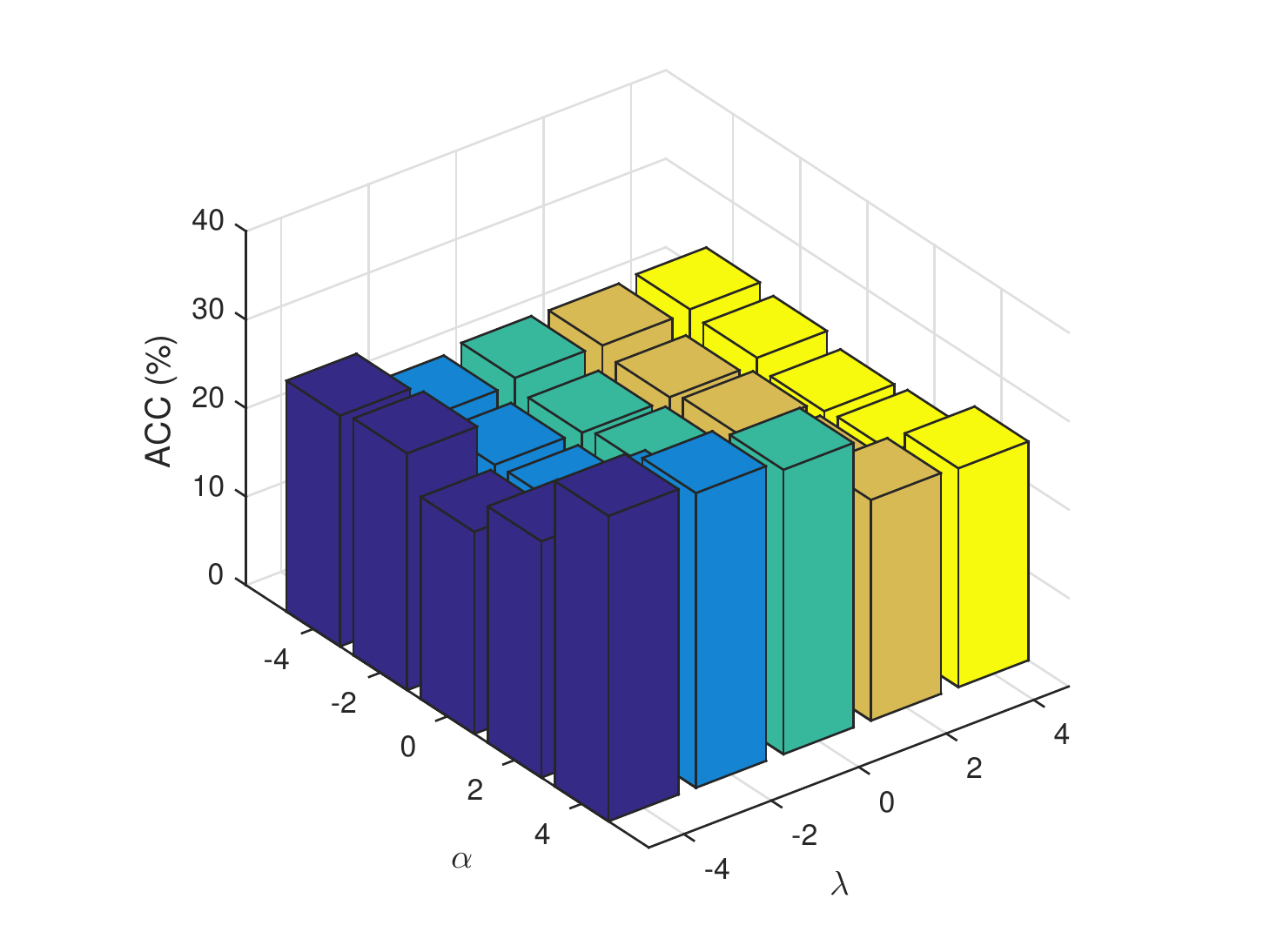}
		\caption{WarpAR10P}
		\label{fig:warpAR10P_acc}
	\end{subfigure}\vspace{1mm}
	\begin{subfigure}[b]{0.32\textwidth}
		\includegraphics[width=\textwidth]{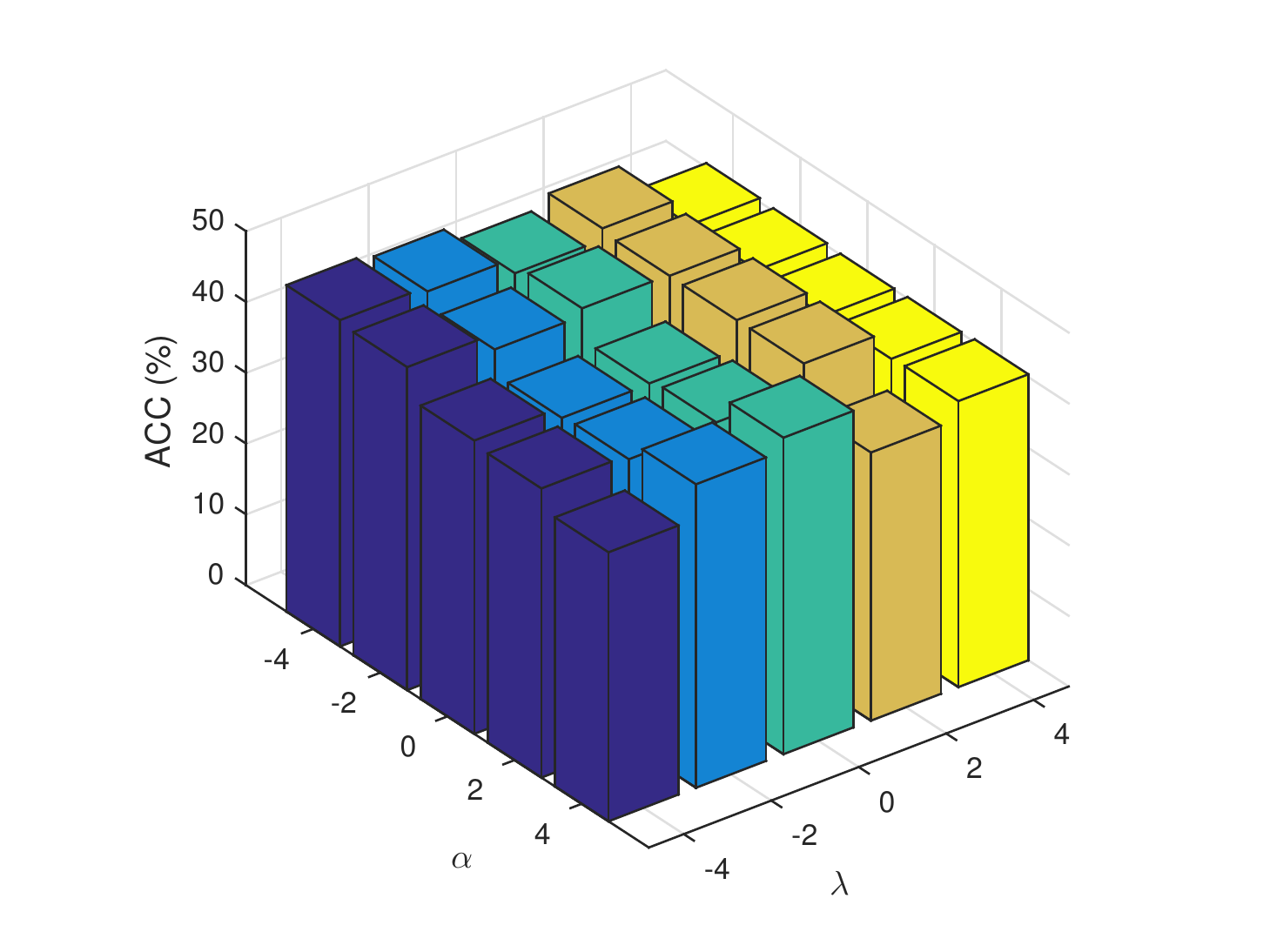}
		\caption{Yale}
		\label{fig:Yale_acc}
	\end{subfigure}\vspace{1mm}
	\caption{Performance of DLUFS in ACC measure using different values of the tuning parameters $\alpha$ and $\beta$ in $\log_{10}$ .}
	\label{fig:sens_acc}
\end{figure}

\begin{figure}[!h]
	\captionsetup[subfigure]{font=scriptsize,labelfont=scriptsize}
	\centering
	\begin{subfigure}[b]{0.32\textwidth}
		\includegraphics[width=\textwidth]{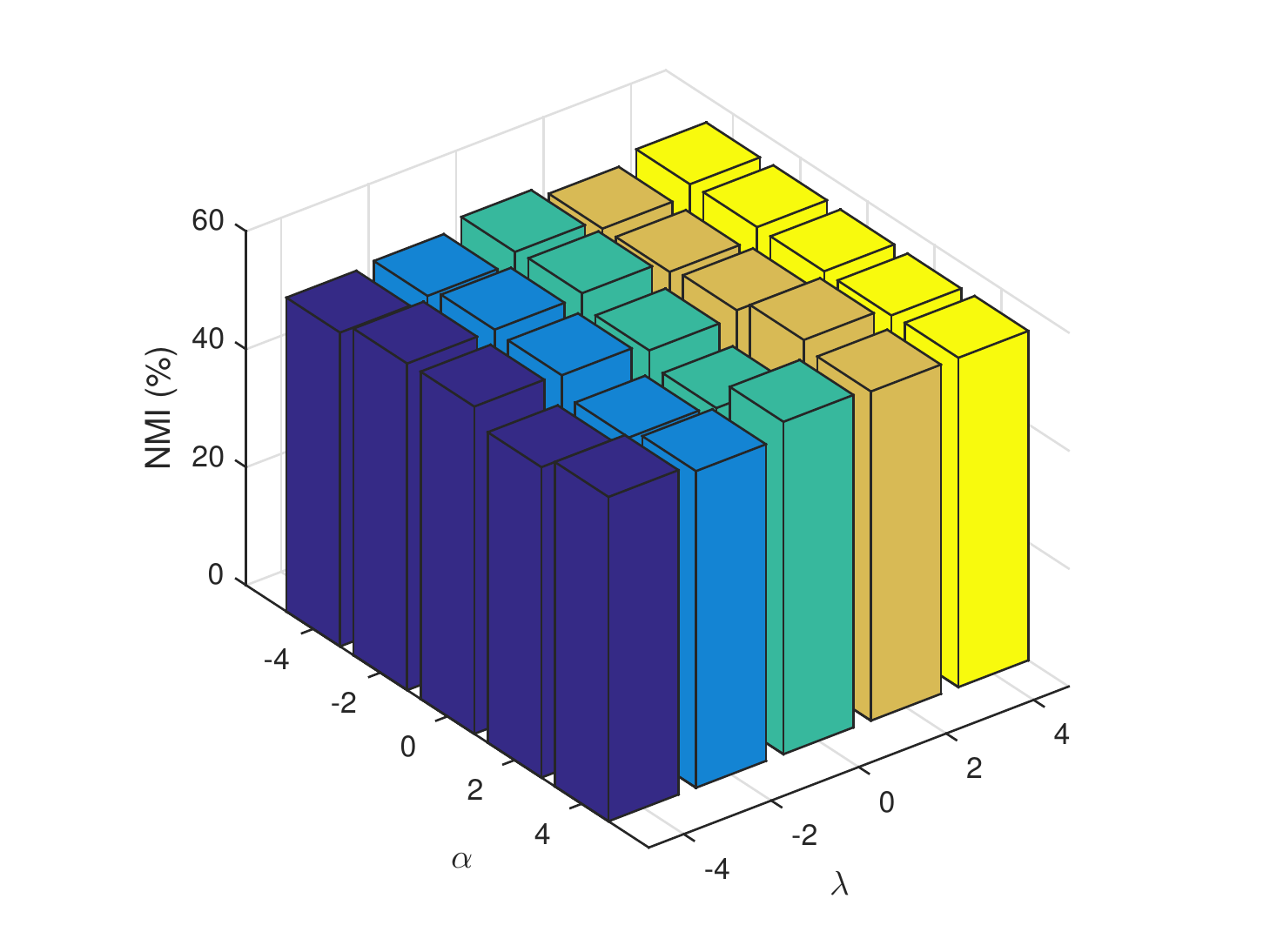}
		\caption{BA}
		\label{fig:BA_NMI}
	\end{subfigure}\vspace{1mm}
	\begin{subfigure}[b]{0.32\textwidth}
		\includegraphics[width=\textwidth]{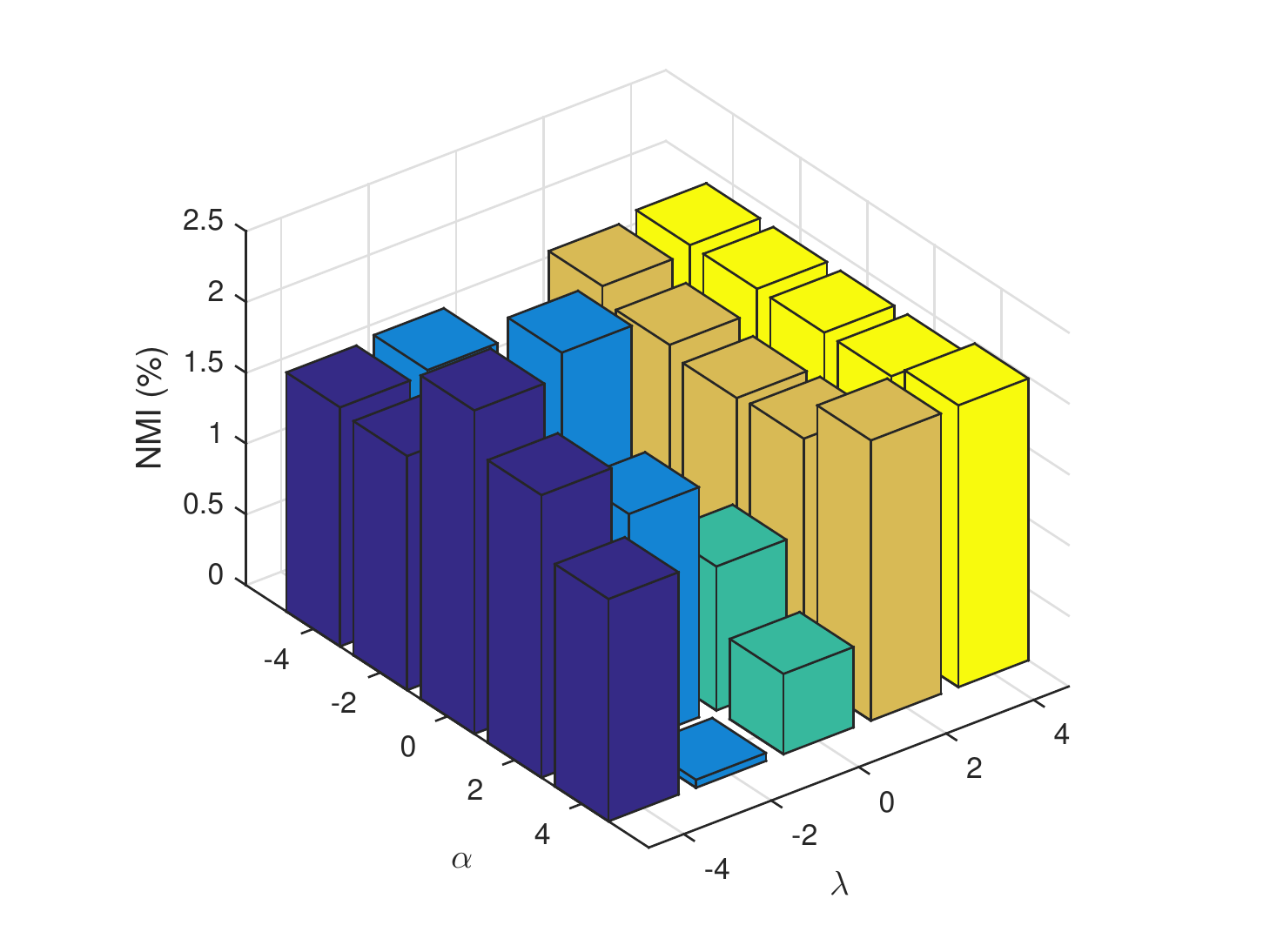}
		\caption{Colon}
		\label{fig:Colon_NMI}
	\end{subfigure}\vspace{1mm}
	\begin{subfigure}[b]{0.32\textwidth}
		\includegraphics[width=\textwidth]{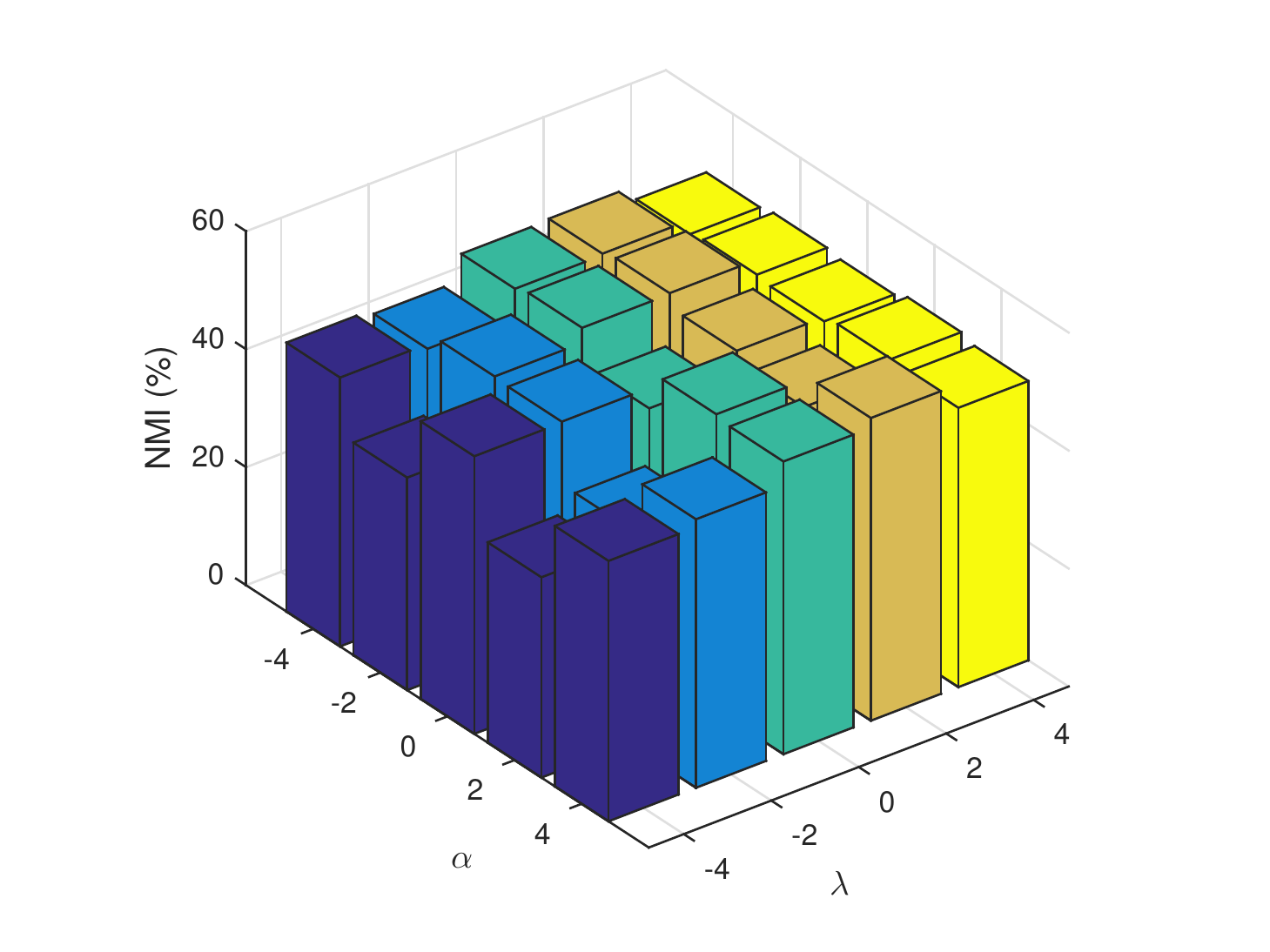}
		\caption{GLIOMA}
		\label{fig:GLIOMA_NMI}
	\end{subfigure}\vspace{1mm}
	\begin{subfigure}[b]{0.32\textwidth}
		\includegraphics[width=\textwidth]{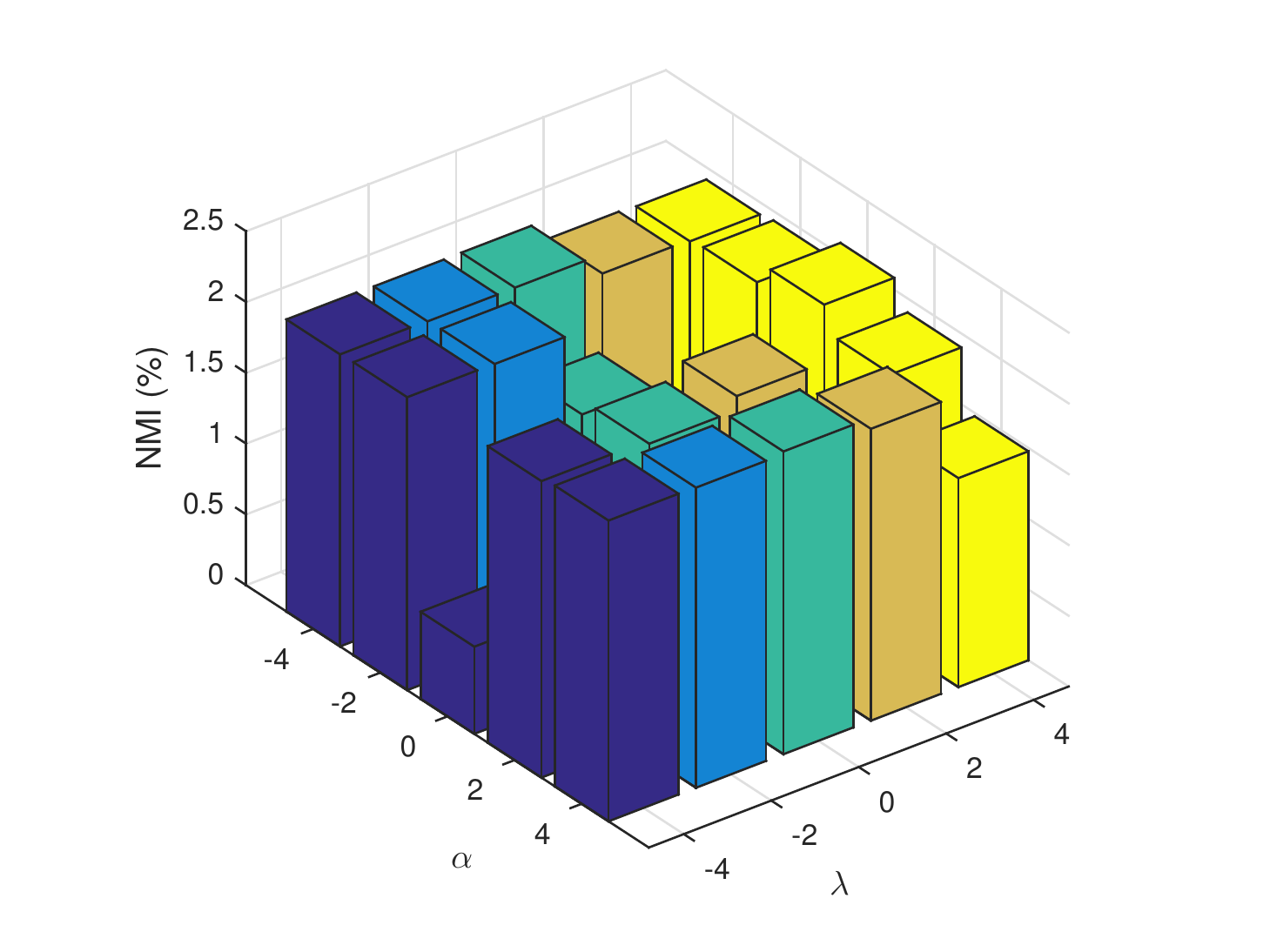}
		\caption{Madelon}
		\label{fig:Madelon_NMI}
	\end{subfigure}\vspace{1mm}
	\begin{subfigure}[b]{0.32\textwidth}
		\includegraphics[width=\textwidth]{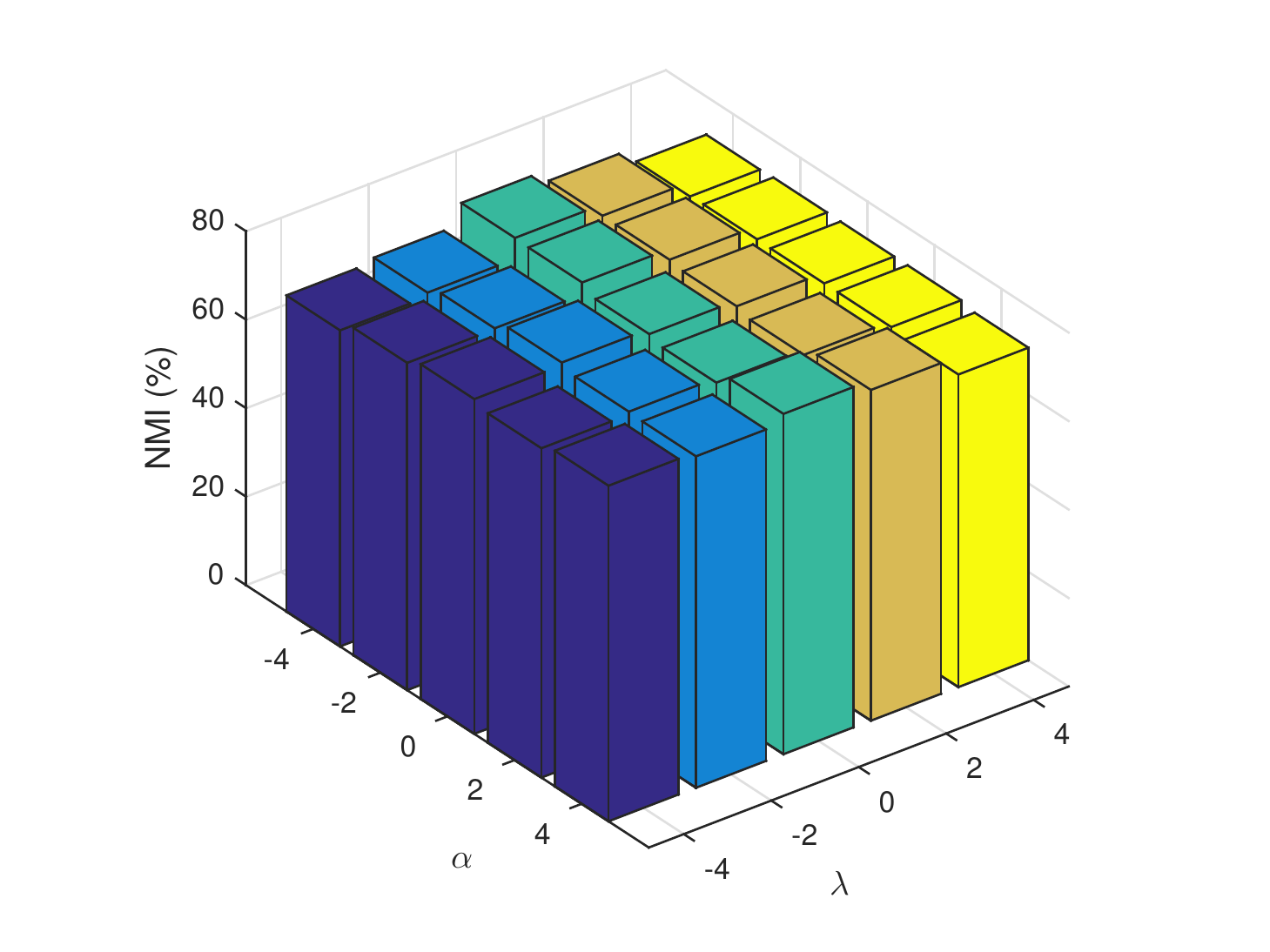}
		\caption{ORL}
		\label{fig:ORL_NMI}
	\end{subfigure}\vspace{1mm}
	\begin{subfigure}[b]{0.32\textwidth}
		\includegraphics[width=\textwidth]{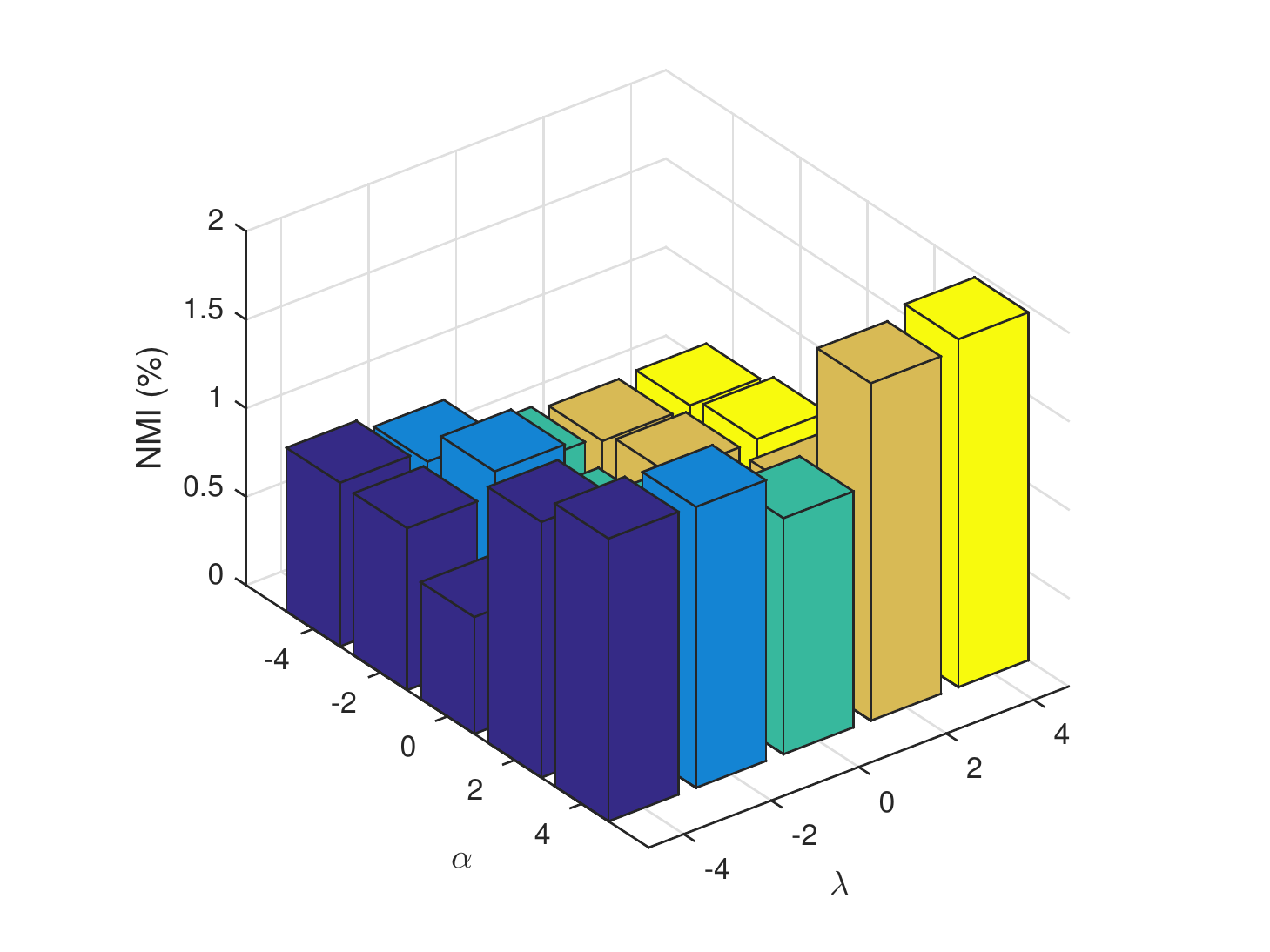}
		\caption{PCMAC}
		\label{fig:PCMAC_NMI}
	\end{subfigure}\vspace{1mm}
	\begin{subfigure}[b]{0.32\textwidth}
		\includegraphics[width=\textwidth]{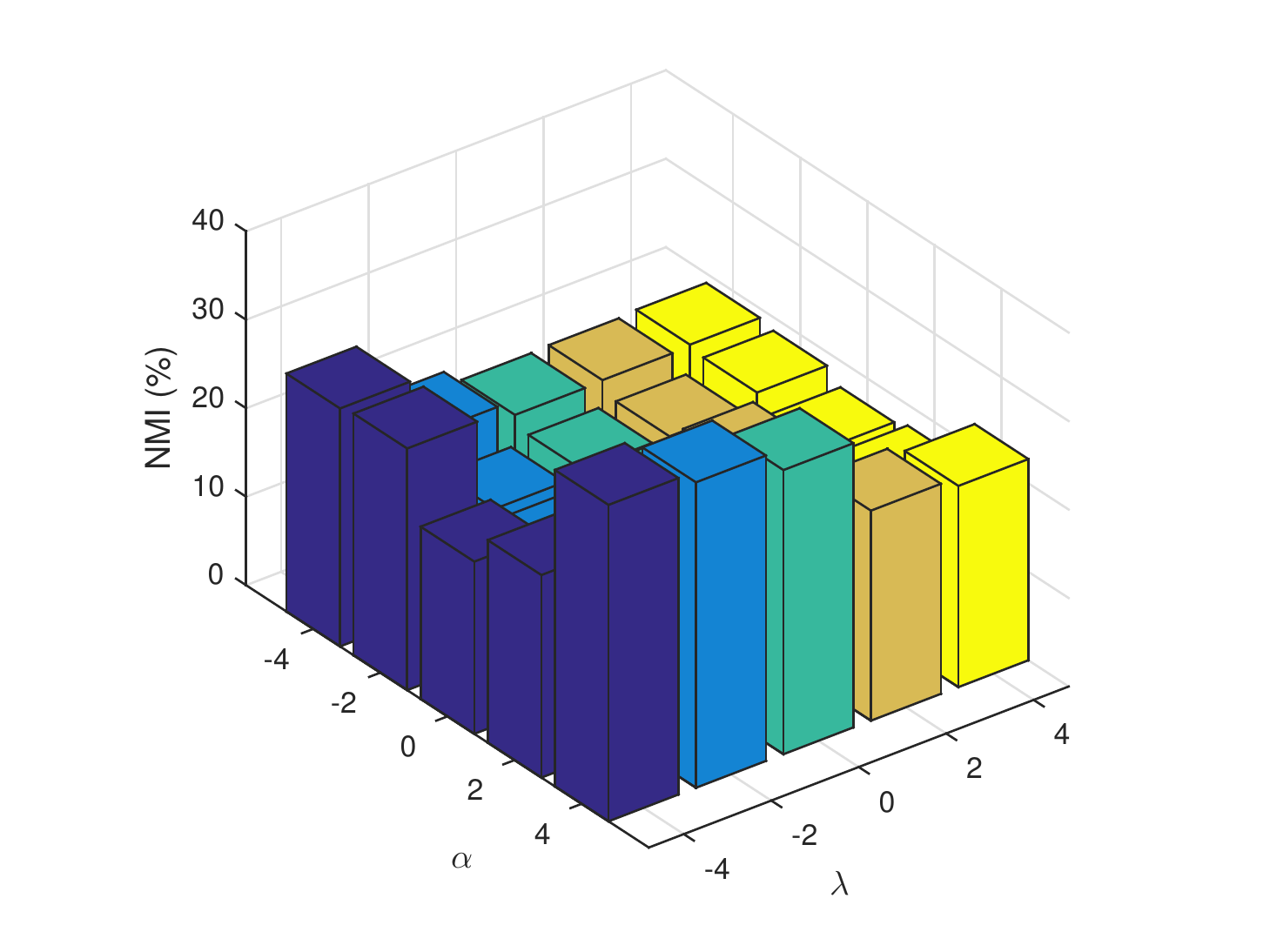}
		\caption{WarpAR10P}
		\label{fig:WarpAR10P_NMI}
	\end{subfigure}\vspace{1mm}
	\begin{subfigure}[b]{0.32\textwidth}
		\includegraphics[width=\textwidth]{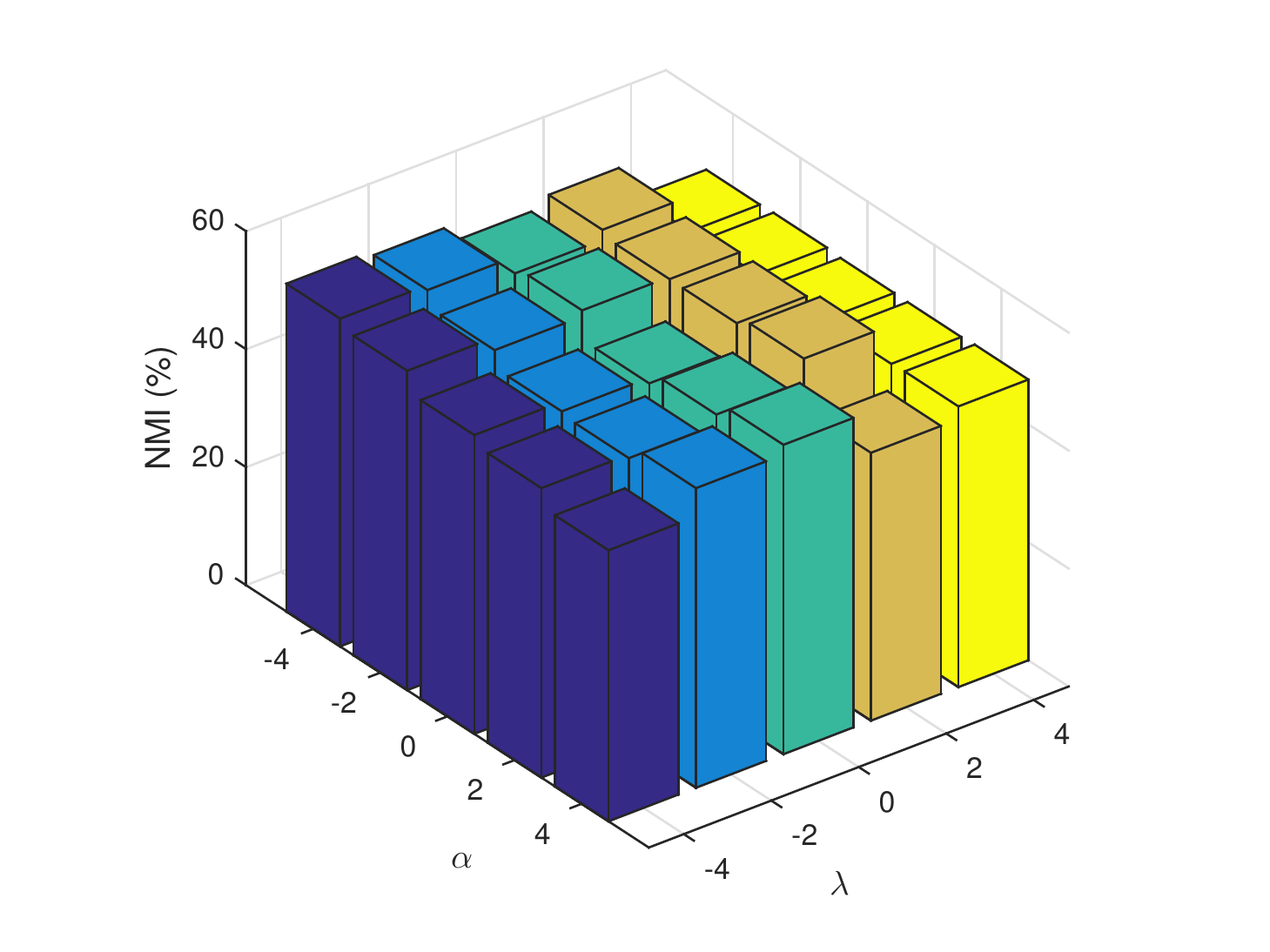}
		\caption{Yale}
		\label{fig:Yale_NMI}
	\end{subfigure}\vspace{1mm}
	\caption{Performance of DLUFS in NMI measure using different values of the tuning parameters $\alpha$ and $\beta$ in $\log_{10}$ .}
	\label{fig:sens_NMI}
\end{figure}

\begin{figure}[!h]
	\captionsetup[subfigure]{font=scriptsize,labelfont=scriptsize}
	\centering
	\begin{subfigure}[b]{0.32\textwidth}
		\begin{tikzpicture}
			\tiny
			\begin{axis}[
				width=\linewidth, 
				grid=major, 
				grid style={dotted}, 
				xlabel=Number of iterations, 
				ylabel=Objective function value,
				xmin=0, xmax=50,
				xtick={0,10,20,30,40,50}
				]
				\addplot[no marks, color=blue, line width=0.75mm]
				coordinates {
					(2,96744.87603)(3,95049.53364)(4,94934.70441)(5,94892.51019)(6,94892.51019)(7,94892.51019)(8,94892.51019)(9,94892.51019)(10,94892.51019)(11,94892.51019)(12,94892.51019)(13,94892.51019)(14,94892.51019)(15,94892.51019)(16,94892.51019)(17,94892.51019)(18,94892.51019)(19,94892.51019)(20,94892.51019)(21,94892.51019)(22,94892.51019)(23,94892.51019)(24,94892.51019)(25,94892.51019)(26,94892.51019)(27,94892.51019)(28,94892.51019)(29,94892.51019)(30,94892.51019)(31,94892.51019)(32,94892.51019)(33,94892.51019)(34,94892.51019)(35,94892.51019)(36,94892.51019)(37,94892.51019)(38,94892.51019)(39,94892.51019)(40,94892.51019)(41,94892.51019)(42,94892.51019)(43,94892.51019)(44,94892.51019)(45,94892.51019)(46,94892.51019)(47,94892.51019)(48,94892.51019)(49,94892.51019)(50,94892.51019)
				};
			\end{axis}
		\end{tikzpicture}
		\vspace{3pt}
		\caption{BA}
		\label{fig:BA_con}
	\end{subfigure}\vspace{1mm}
	\begin{subfigure}[b]{0.32\textwidth}
		\begin{tikzpicture}
			\tiny
			\begin{axis}[
				width=\linewidth, 
				grid=major, 
				grid style={dotted}, 
				xlabel=Number of iterations, 
				ylabel=Objective function value,
				xmin=0, xmax=50,
				xtick={0,10,20,30,40,50}
				]
				\addplot[no marks, color=blue, line width=0.75mm]
				coordinates {
					(2,3955.018555)(3,3863.884079)(4,3852.828131)(5,3850.206142)(6,3850.206142)(7,3850.206142)(8,3850.206142)(9,3850.206142)(10,3850.206142)(11,3850.206142)(12,3850.206142)(13,3850.206142)(14,3850.206142)(15,3850.206142)(16,3850.206142)(17,3850.206142)(18,3850.206142)(19,3850.206142)(20,3850.206142)(21,3850.206142)(22,3850.206142)(23,3850.206142)(24,3850.206142)(25,3850.206142)(26,3850.206142)(27,3850.206142)(28,3850.206142)(29,3850.206142)(30,3850.206142)(31,3850.206142)(32,3850.206142)(33,3850.206142)(34,3850.206142)(35,3850.206142)(36,3850.206142)(37,3850.206142)(38,3850.206142)(39,3850.206142)(40,3850.206142)(41,3850.206142)(42,3850.206142)(43,3850.206142)(44,3850.206142)(45,3850.206142)(46,3850.206142)(47,3850.206142)(48,3850.206142)(49,3850.206142)(50,3850.206142)
				};
			\end{axis}
		\end{tikzpicture}
		\vspace{3pt}
		\caption{Colon}
		\label{fig:Colon_con}
	\end{subfigure}\vspace{1mm}
	\begin{subfigure}[b]{0.32\textwidth}
		\begin{tikzpicture}
			\tiny
			\begin{axis}[
				width=\linewidth, 
				grid=major, 
				grid style={dotted}, 
				xlabel=Number of iterations, 
				ylabel=Objective function value,
				xmin=0, xmax=50,
				xtick={0,10,20,30,40,50}
				]
				\addplot[no marks, color=blue, line width=0.75mm]
				coordinates {
					(2,505.8117915)(3,472.3759599)(4,444.0187058)(5,420.2339813)(6,399.9991713)(7,382.5690779)(8,367.4118489)(9,354.1331197)(10,342.430181)(11,332.0644896)(12,322.8443023)(13,314.6131155)(14,307.2416199)(15,300.6218981)(16,294.6631152)(17,289.2882472)(18,284.4315493)(19,280.0365694)(20,276.0545695)(21,272.4432594)(22,269.165771)(23,266.1898217)(24,263.487027)(25,261.0323331)(26,258.8035443)(27,256.7809283)(28,254.9468851)(29,253.285667)(30,251.783142)(31,250.4265915)(32,249.2045383)(33,248.1065986)(34,247.1233547)(35,246.246245)(36,245.4674688)(37,244.7799032)(38,244.1770312)(39,243.6528782)(40,243.2019568)(41,242.8192179)(42,242.5000076)(43,242.2400296)(44,242.0353113)(45,242.0353113)(46,242.0353113)(47,242.0353113)(48,242.0353113)(49,242.0353113)(50,242.0353113)
				};
			\end{axis}
		\end{tikzpicture}
		\vspace{3pt}
		\caption{GLIOMA}
		\label{fig:GLIOMA_con}
	\end{subfigure}\vspace{1mm}
	\begin{subfigure}[b]{0.32\textwidth}
		\begin{tikzpicture}
			\tiny
			\begin{axis}[
				width=\linewidth, 
				grid=major, 
				grid style={dotted}, 
				xlabel=Number of iterations, 
				ylabel=Objective function value,
				xmin=0, xmax=50,
				xtick={0,10,20,30,40,50}
				]
				\addplot[no marks, color=blue, line width=0.75mm]
				coordinates {
					(2,1040778.027)(3,1035802.603)(4,1035278.181)(5,1035278.181)(6,1035278.181)(7,1035278.181)(8,1035278.181)(9,1035278.181)(10,1035278.181)(11,1035278.181)(12,1035278.181)(13,1035278.181)(14,1035278.181)(15,1035278.181)(16,1035278.181)(17,1035278.181)(18,1035278.181)(19,1035278.181)(20,1035278.181)(21,1035278.181)(22,1035278.181)(23,1035278.181)(24,1035278.181)(25,1035278.181)(26,1035278.181)(27,1035278.181)(28,1035278.181)(29,1035278.181)(30,1035278.181)(31,1035278.181)(32,1035278.181)(33,1035278.181)(34,1035278.181)(35,1035278.181)(36,1035278.181)(37,1035278.181)(38,1035278.181)(39,1035278.181)(40,1035278.181)(41,1035278.181)(42,1035278.181)(43,1035278.181)(44,1035278.181)(45,1035278.181)(46,1035278.181)(47,1035278.181)(48,1035278.181)(49,1035278.181)(50,1035278.181)
				};
			\end{axis}
		\end{tikzpicture}
		\vspace{3pt}
		\caption{Madelon}
		\label{fig:Madelon_con}
	\end{subfigure}\vspace{1mm}
	\begin{subfigure}[b]{0.32\textwidth}
		\begin{tikzpicture}
			\tiny
			\begin{axis}[
				width=\linewidth, 
				grid=major, 
				grid style={dotted}, 
				xlabel=Number of iterations, 
				ylabel=Objective function value,
				xmin=0, xmax=50,
				xtick={0,10,20,30,40,50}
				]
				\addplot[no marks, color=blue, line width=0.75mm]
				coordinates {
					(2,57921.63912)(3,53960.78434)(4,52707.06068)(5,52162.0714)(6,51755.78284)(7,51615.05096)(8,51540.54576)(9,51482.44818)(10,51430.6934)(11,51384.24551)(12,51384.24551)(13,51384.24551)(14,51384.24551)(15,51384.24551)(16,51384.24551)(17,51384.24551)(18,51384.24551)(19,51384.24551)(20,51384.24551)(21,51384.24551)(22,51384.24551)(23,51384.24551)(24,51384.24551)(25,51384.24551)(26,51384.24551)(27,51384.24551)(28,51384.24551)(29,51384.24551)(30,51384.24551)(31,51384.24551)(32,51384.24551)(33,51384.24551)(34,51384.24551)(35,51384.24551)(36,51384.24551)(37,51384.24551)(38,51384.24551)(39,51384.24551)(40,51384.24551)(41,51384.24551)(42,51384.24551)(43,51384.24551)(44,51384.24551)(45,51384.24551)(46,51384.24551)(47,51384.24551)(48,51384.24551)(49,51384.24551)(50,51384.24551)
				};
			\end{axis}
		\end{tikzpicture}
		\vspace{3pt}
		\caption{ORL}
		\label{fig:ORL_con}
	\end{subfigure}\vspace{1mm}
	\begin{subfigure}[b]{0.32\textwidth}
		\begin{tikzpicture}
			\tiny
			\begin{axis}[
				width=\linewidth, 
				grid=major, 
				grid style={dotted}, 
				xlabel=Number of iterations, 
				ylabel=Objective function value,
				xmin=0, xmax=50,
				xtick={0,10,20,30,40,50}
				]
				\addplot[no marks, color=blue, line width=0.75mm]
				coordinates {
					(2,4775555.289)(3,4620788.829)(4,4602732.681)(5,4596434.818)(6,4593497.268)(7,4593497.268)(8,4593497.268)(9,4593497.268)(10,4593497.268)(11,4593497.268)(12,4593497.268)(13,4593497.268)(14,4593497.268)(15,4593497.268)(16,4593497.268)(17,4593497.268)(18,4593497.268)(19,4593497.268)(20,4593497.268)(21,4593497.268)(22,4593497.268)(23,4593497.268)(24,4593497.268)(25,4593497.268)(26,4593497.268)(27,4593497.268)(28,4593497.268)(29,4593497.268)(30,4593497.268)(31,4593497.268)(32,4593497.268)(33,4593497.268)(34,4593497.268)(35,4593497.268)(36,4593497.268)(37,4593497.268)(38,4593497.268)(39,4593497.268)(40,4593497.268)(41,4593497.268)(42,4593497.268)(43,4593497.268)(44,4593497.268)(45,4593497.268)(46,4593497.268)(47,4593497.268)(48,4593497.268)(49,4593497.268)(50,4593497.268)
				};
			\end{axis}
		\end{tikzpicture}
		\vspace{3pt}
		\caption{PCMAC}
		\label{fig:PCMAC_con}
	\end{subfigure}\vspace{1mm}
	\begin{subfigure}[b]{0.32\textwidth}
		\begin{tikzpicture}
			\tiny
			\begin{axis}[
				width=\linewidth, 
				grid=major, 
				grid style={dotted}, 
				xlabel=Number of iterations, 
				ylabel=Objective function value,
				xmin=0, xmax=50,
				xtick={0,10,20,30,40,50}
				]
				\addplot[no marks, color=blue, line width=0.75mm]
				coordinates {
					(2,14790.36743)(3,14167.89796)(4,14122.22752)(5,14107.29507)(6,14097.87352)(7,14097.87352)(8,14097.87352)(9,14097.87352)(10,14097.87352)(11,14097.87352)(12,14097.87352)(13,14097.87352)(14,14097.87352)(15,14097.87352)(16,14097.87352)(17,14097.87352)(18,14097.87352)(19,14097.87352)(20,14097.87352)(21,14097.87352)(22,14097.87352)(23,14097.87352)(24,14097.87352)(25,14097.87352)(26,14097.87352)(27,14097.87352)(28,14097.87352)(29,14097.87352)(30,14097.87352)(31,14097.87352)(32,14097.87352)(33,14097.87352)(34,14097.87352)(35,14097.87352)(36,14097.87352)(37,14097.87352)(38,14097.87352)(39,14097.87352)(40,14097.87352)(41,14097.87352)(42,14097.87352)(43,14097.87352)(44,14097.87352)(45,14097.87352)(46,14097.87352)(47,14097.87352)(48,14097.87352)(49,14097.87352)(50,14097.87352)
				};
			\end{axis}
		\end{tikzpicture}
		\vspace{3pt}
		\caption{WarpAR10P}
		\label{fig:warpAR10P_con}
	\end{subfigure}\vspace{1mm}
	\begin{subfigure}[b]{0.32\textwidth}
		\begin{tikzpicture}
			\tiny
			\begin{axis}[
				width=\linewidth, 
				grid=major, 
				grid style={dotted}, 
				xlabel=Number of iterations, 
				ylabel=Objective function value,
				xmin=0, xmax=50,
				xtick={0,10,20,30,40,50}
				]
				\addplot[no marks, color=blue, line width=0.75mm]
				coordinates {
					(2,16444.09382)(3,15667.20471)(4,15429.31043)(5,15383.24556)(6,15365.71047)(7,15353.6936)(8,15353.6936)(9,15353.6936)(10,15353.6936)(11,15353.6936)(12,15353.6936)(13,15353.6936)(14,15353.6936)(15,15353.6936)(16,15353.6936)(17,15353.6936)(18,15353.6936)(19,15353.6936)(20,15353.6936)(21,15353.6936)(22,15353.6936)(23,15353.6936)(24,15353.6936)(25,15353.6936)(26,15353.6936)(27,15353.6936)(28,15353.6936)(29,15353.6936)(30,15353.6936)(31,15353.6936)(32,15353.6936)(33,15353.6936)(34,15353.6936)(35,15353.6936)(36,15353.6936)(37,15353.6936)(38,15353.6936)(39,15353.6936)(40,15353.6936)(41,15353.6936)(42,15353.6936)(43,15353.6936)(44,15353.6936)(45,15353.6936)(46,15353.6936)(47,15353.6936)(48,15353.6936)(49,15353.6936)(50,15353.6936)
				};
			\end{axis}
		\end{tikzpicture}
		\vspace{3pt}
		\caption{Yale}
		\label{fig:Yale_con}
	\end{subfigure}\vspace{1mm}
	
	\caption{Convergence curve of DLUFS on different datasets.}
	\label{fig:convergence}
\end{figure}
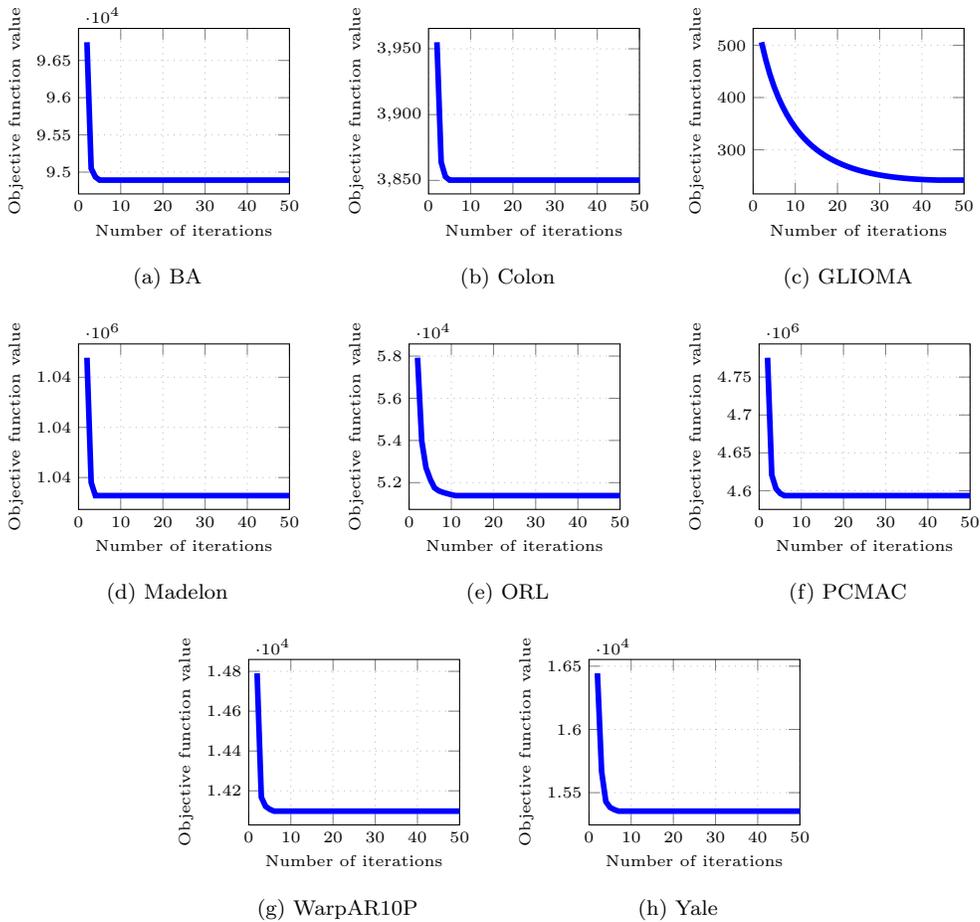
  
\subsection{Experimental results}
We demonstrate  the performance of the proposed method, DLUFS, by comparing to benchmark UFS methods defined in the earlier subsection. \mbox{Table \ref{tab:acc_res} and \ref{tab:nmi_res}} represent the obtained results based on ACC and NMI (mean $\pm$ STD). The bold numbers represent the best attained results. We used the underlined numbers to indicate the second-best results. We summarize the main findings from the obtained results of Table \mbox{\ref{tab:acc_res} and Table \ref{tab:nmi_res}} in the following,
\begin{itemize}
	\item In most cases, DLUFS outperforms the Baseline, which would lead to the superiority of selecting features over learning on all features. Therefore, feature selection process improves the efficiency (i.e. ease of computation) and performance(i.e. better results).
	\item  DLUFS obtain good results as the best or the second-best (after Baseline) aligned with the good performance of LDSSL, SRFS which could be due to data reconstruction.
	\item Our method achieves good performance in most datasets. It can be attributed to low-rank representation as SRFS.
\end{itemize} 

Furthermore, the performance of the proposed method is demonstrated by selecting various number of features from $50$ to $300$. \mbox{Fig. \ref{fig:acc_linechart1} and Fig. \ref{fig:NMI_linechart1}} show the results for different numbers of selected features, ignoring LS, MCFS, SPFS, and UDFS due to weak results.
Obviously, DLUFS achieves the best performance compared to the competitors even when the number of selected features is varied.

\subsection{Parameter sensitivity and convergence study}
First, the sensitivity of the parameters is investigated in DLUFS. The main parameters of the method are $\alpha$ and $\lambda$, by considering the objective function of DLUFS (Eq. \eqref{eq:14}). We explore the set of candidate values for tuning parameters $\alpha$ and $\lambda$ from $\{10^{-4}, 10^{-2}, 1, 10^2, 10^4\}$. The datasets of Table \ref{tb_datasets} are used to investigate the affect of variations in the main tuning parameters.  We report the obtained results of DLUFS by considering ACC and NMI measures. Fig. \ref{fig:sens_acc} shows ACC values based on different settings of   $\alpha$  and $\lambda$ parameters, and Fig. \ref{fig:sens_NMI} represents NMI values.  The results indicate that there exists a slight sensitivity to these main parameters in the performance of DLUFS.

Next, the convergence behavior of the DLUFS algorithm is experimentally demonstrated on the datasets of Table \ref{tb_datasets}. The convergence speed of the objective function is depicted versus the number of iterations in \mbox{Fig. \ref{fig:convergence}}.  These findings from \mbox{Fig. \ref{fig:convergence}} reveal the efficiency of the proposed algorithm based on the convergence speed.

\section{Conclusion}\label{sec:concl}
In this work, we proposed a new unsupervised feature selection method based on dictionary learning to learn a new data representation in a basis space. We devise a low-rank constraint on the basis matrix to preserve the feature correlation along with subspace learning. Spectral analysis was employed to consider the sample similarities in the learned data representation matrix. Moreover, an $\ell_{2,1}$-norm regularization was applied in our primary objective function to discard uninformative features.  We presented a unified framework based on the important characteristics to select features.
An efficient numerical algorithm is introduced for the proposed method.  The performance of the proposed method was investigated through state-of-the-art methods by the aid of a variety of standard datasets. The attained results verified the strength of the proposed approach in terms of accuracy and speed of convergence.
\\
\\
\section*{}
\bibliographystyle{model5-names}
\biboptions{authoryear}
\bibliography{DLUFS.bib}
\end{document}